\documentclass[a4paper,fleqn]{cas-sc}

\usepackage{apacite}
\usepackage[authoryear]{natbib}

\usepackage{url}
\usepackage{multirow}
\usepackage{listings}
\usepackage{algorithm}
\usepackage{algpseudocode}
\usepackage{booktabs}
\usepackage[most]{tcolorbox}
\usepackage{subcaption}
\usepackage{amssymb}
\usepackage{amsmath}
\usepackage{graphicx} 

\newtcolorbox{grayquote}{
    colback=gray!10,
    colframe=white,
    boxrule=0pt,
    left=5pt,
    right=5pt,
    top=5pt,
    bottom=5pt
}

\def\tsc#1{\csdef{#1}{\textsc{\lowercase{#1}}\xspace}}
\tsc{WGM}
\tsc{QE}
\tsc{EP}
\tsc{PMS}
\tsc{BEC}
\tsc{DE}

\begin{document}
\let\WriteBookmarks\relax
\def\floatpagepagefraction{1}
\def\textpagefraction{.001}

\shorttitle{LLM-Confidence Reranker}

\shortauthors{Song et~al.}

\title[mode = title]{LLM-Confidence Reranker: A Training-Free Approach for Enhancing Retrieval-Augmented Generation Systems} 

\author[dut]{Zhipeng~Song}
[orcid=0009-0009-6249-1988]
\cormark[1]
\ead{songzhipeng@mail.dlut.edu.cn}

\author[ldu]{Xiangyu~Kong}
[orcid=0000-0003-1940-8674]
\cormark[1]
\ead{xiangyukong@liaodongu.edu.cn}

\author[ltu]{Xinrui~Bao}
[orcid=0009-0000-7916-2122]
\ead{4724200573@stu.lntu.edu.cn}

\author[dlou]{Yizhi~Zhou}
[orcid=0000-0002-6761-5953]
\ead{zhouyizhi@dlou.edu.cn}

\author[dut,qhu]{Jiulong~Jiao}
[orcid=0009-0001-9852-7999]
\ead{jiaojiulong@mail.dlut.edu.cn}

\author[tnu]{Sitong~Liu}
[orcid=0009-0000-4728-7078]
\ead{liusitong@stu.tjnu.edu.cn}

\author[tencentdl]{Yuhang~Zhou}
[orcid=0009-0007-3489-0549]
\ead{ginozhou@tencent.com}

\author[dut]{Heng~Qi}
[orcid=0000-0002-8770-3934]
\cormark[2]
\ead{hengqi@dlut.edu.cn}

\cortext[equal]{The two authors contribute equally to this work and should be regarded as co-first authors.}
\cortext[cor1]{Corresponding author.}






\affiliation[dut]{organization={School of Computer Science and Technology, Dalian~University~of~Technology},
    addressline={No.2 Linggong Road, Ganjingzi District},
    city={Dalian},
    postcode={116024}, 
    country={China}}

\affiliation[ldu]{organization={School of Information Engineering, Liaodong~University},
    addressline={No.116 Linjiang Back Street, Zhenan District},
    city={Dandong},
    postcode={118001}, 
    country={China}}

\affiliation[dlou]{organization={School of Information Engineering, Dalian Ocean University},
    addressline={No. 2-52, Heishijiao Street, Shahekou District},
    city={Dalian},
    postcode={116023}, 
    country={China}}

\affiliation[qhu]{
organization={Information Technology Center, Qinghai~University},
    addressline={251 Ningda Road, Chengbei District},
    city={Xining},
    postcode={810016}, 
    country={China}}

\affiliation[ltu]{
organization={School of Electronic and Information Engineering, Liaoning~Technical~University},
    addressline={188 Longwan South Street, Sijiatun District},
    city={Huludao},
    postcode={125105}, 
    country={China}}

\affiliation[tnu]{organization={School of Artificial Intelligence,
Tianjin~Normal~University},
    addressline={393 Binshui West Road, Xiqing District},
    city={Tianjin},
    postcode={300387}, 
    country={China}}

\affiliation[tencentdl]{organization={Tencent (Dalian Northern Interactive Entertainment Technology Co., Ltd.)},
    addressline={21/F, Tencent Building, No. 26 Jingxian St, Ganjingzi District},
    city={Dalian},
    postcode={116085}, 
    country={China}}

\title [mode = title]{LLM-Confidence Reranker: A Training-Free Approach for Enhancing Retrieval-Augmented Generation Systems} 

\begin{abstract}
Large language models (LLMs) have revolutionized natural language processing, yet hallucinations in knowledge-intensive tasks remain a critical challenge. Retrieval-augmented generation (RAG) addresses this by integrating external knowledge, but its efficacy depends on accurate document retrieval and ranking. Although existing rerankers demonstrate effectiveness, they frequently necessitate specialized training, impose substantial computational expenses, and fail to fully exploit the semantic capabilities of LLMs, particularly their inherent confidence signals. We propose the LLM-Confidence Reranker (LCR), a training-free, plug-and-play algorithm that enhances reranking in RAG systems by leveraging black-box LLM confidence derived from Maximum Semantic Cluster Proportion (MSCP). LCR employs a two-stage process: confidence assessment via multinomial sampling and clustering, followed by binning and multi-level sorting based on query and document confidence thresholds. This approach prioritizes relevant documents while preserving original rankings for high-confidence queries, ensuring robustness.
Evaluated on BEIR and TREC benchmarks with BM25 and Contriever retrievers, LCR—using only 7--9B-parameter pre-trained LLMs—consistently improves NDCG@5 by up to 20.6\% across pre-trained LLM and fine-tuned Transformer rerankers, without degradation. Ablation studies validate the hypothesis that LLM confidence positively correlates with document relevance, elucidating LCR's mechanism.
LCR offers computational efficiency, parallelism for scalability, and broad compatibility, mitigating hallucinations in applications like medical diagnosis.
\end{abstract}


\maketitle



\section{Introduction}
\label{sec:intro}

In recent years, large language models (LLMs) have achieved remarkable progress in natural language processing (NLP). They excel in tasks such as text generation and question answering. Leveraging their advanced capabilities in language understanding and generation, LLMs have significantly enhanced task efficiency and user experience across a wide range of applications \citep{guo2025deepseek}. However, a persistent challenge remains: ``hallucination''---the generation of plausible yet factually inaccurate or entirely fabricated content. This issue poses a significant barrier in knowledge-intensive applications, such as medical diagnosis, where factual accuracy is paramount \citep{ji2023survey, huang2025survey, luo2026rallrec, plonka2025comparative, chi2026generalized, li2025llm}.

To mitigate hallucination, retrieval-augmented generation (RAG) has emerged as a promising approach. By integrating external knowledge retrieval with the generation process, RAG provides LLMs with a factual foundation. This improves the accuracy of generated outputs \citep{lewis2020retrieval, yu2022retrieval, yao2026collaborative, ren2025retrieval}. Hallucinations often stem from low confidence in internal knowledge, which can be addressed by external documents that boost model confidence. Nevertheless, the success of RAG hinges on the relevance of retrieved documents. Poorly ranked documents undermine the quality of subsequent generation. This limits the framework’s effectiveness \citep{liu2024lost}.

Reranking has gained prominence as a vital step in enhancing retrieval outcomes within RAG systems. By reordering initially retrieved documents, reranking techniques improve document relevance and provide higher-quality inputs for generation \citep{yu2024rankrag, salemi2024evaluating, xu2024list, ren2025large}.
Although existing reranking methods have advanced retrieval performance, they encounter notable limitations. Many depend on proprietary models trained on specific datasets, leading to poor generalization across domains and high computational costs \citep{ma2023zero, sun2023is, wang2023well, dong2024don}---constraints that severely restrict their practical applicability.
In contrast, pre-trained LLM-based rerankers provide strong generalization and no additional training costs compared to trained models; however, existing LLM-based rerankers primarily assess query-document relevance but underutilize LLMs' strengths in semantic understanding, particularly by overlooking confidence signals from generation processes. 
Furthermore, even state-of-the-art rerankers demonstrate suboptimal performance, with their Normalized Discounted Cumulative Gain (NDCG) scores often falling below 0.8 on benchmarks like BEIR \citep{kamalloo2024resources}. This underscores the necessity for continued innovation in reranking methodologies.

\begin{figure}[pos=htbp]
\centering
\includegraphics[width=\linewidth]{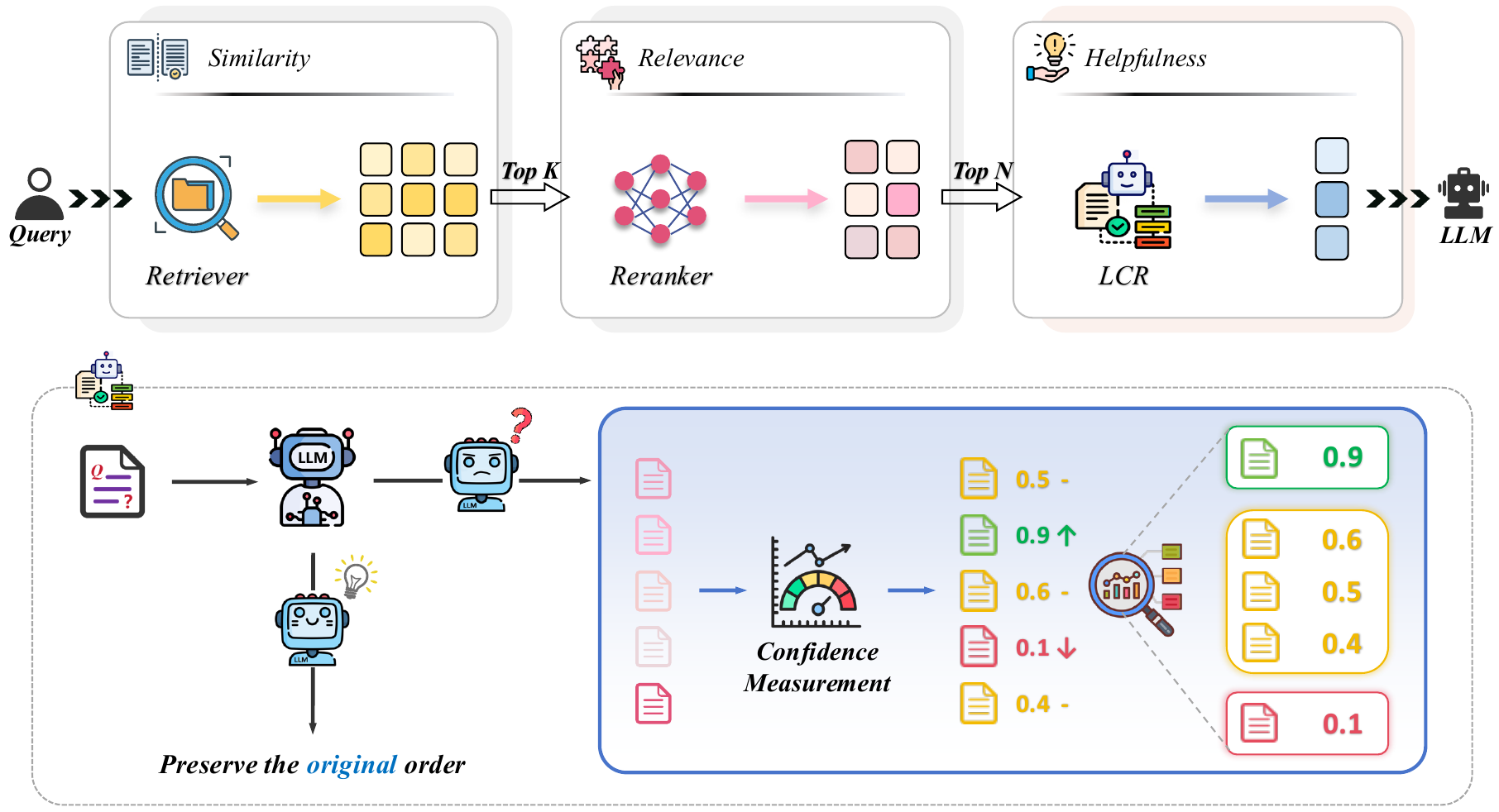}
\caption{\textbf{Illustration of LLM-Confidence Reranker (LCR) algorithm.} LCR leverages LLM confidence signals derived from document helpfulness to enhance ranking, distinct from traditional relevance or similarity measures.}
\label{fig:lcr}
\end{figure}

The main contributions of this study are as follows:
\begin{enumerate}
    \item We propose a novel zero-shot reranking method LLM-Confidence Reranker (LCR) based on black-box LLM confidence signals, which quantifies confidence through the Maximum Semantic Cluster Proportion (MSCP). This approach fully leverages the LLM's semantic understanding and question-answering capabilities to enhance document ranking accuracy in RAG systems.
    \item LCR exhibits high adaptability and generalization, functioning as a standalone reranker or seamlessly integrating after existing rerankers to further refine ranking results. The approach relies solely on lightweight pre-trained LLMs with 7-9B parameters, requiring no training or internal parameter access, ensuring high computational efficiency and low cost, and achieving consistent performance improvements across diverse retrievers and rerankers on the BEIR and TREC benchmarks, achieving NDCG@5 gains of up to 20.6\%.
    \item This method employs a pointwise independent scoring mechanism, supporting parallel processing and scalability to large-scale documents.
    \item Through experiments, we validate the hypothesis that LLM confidence in generated responses positively correlates with document relevance, elucidating the effective mechanism of LCR and providing a theoretical foundation for mitigating hallucinations in knowledge-intensive tasks.
\end{enumerate}

The remainder of the paper is organized as follows: Section~\ref{sec:related-work} reviews the related literature. 
Section~\ref{sec:methodology} elucidates the confidence modeling techniques and delineates the LCR algorithm. 
Section~\ref{sec:experiments} details the experimental evaluations, including the setup, results, and analyses of the impacts of retrievers, models, hyperparameters, and confidence quantification methods, as well as the underlying mechanism. 
Section~\ref{sec:conclusion} summarizes the findings and outlines future research directions.

\section{Related Work}
\label{sec:related-work}

\subsection{Uncertainty Quantification Methods for LLMs}
Uncertainty quantification methods can be categorized into two main types depending on whether model training is necessary: training-required methods and training-free methods. Training-required methods improve the uncertainty expression capabilities of large language models (LLMs) by employing fine-tuning or other learning processes. In contrast, training-free methods directly leverage the model's pre-existing capabilities to extract uncertainty information.

\subsubsection{Training-Required Methods}
Training-required methods improve LLMs' ability to accurately express uncertainty through model adjustment or optimization, primarily following three technical approaches:

\paragraph{Supervised Fine-tuning} Supervised fine-tuning aims to enhance LLMs' uncertainty expression through training. For example, \citet{lin2022teaching} enabled models to express uncertainty in natural language via fine-tuning. A key development is the P(IK) method, proposed by \citet{kadavath2022language}, which quantifies uncertainty by training language models to predict their own probability of correctly answering questions. To enhance the model's transparency and explainability regarding uncertainty, \citet{yang2024alignment} further explored the possibility of embedding explicit confidence in the model's output. This was achieved by training the model to use language such as ``I'm absolutely certain'' or express numerical confidence (e.g., ``about 90\% confident''), allowing this confidence to be explicitly presented to the user. In subsequent research, scholars began to focus on whether the decisiveness of language in model generations faithfully reflects their intrinsic confidence. \citet{yona2024can} formalized faithful response uncertainty based on the gap between linguistic decisiveness and intrinsic confidence, evaluating models' ability to faithfully express uncertainty in words.\citet{manggala2025qa} systematically explores the discrepancy between confidence scores and actual accuracy, proposing QA-calibration as a generalized, principled calibration notion, which provides a reference direction for improving uncertainty expression in generative QA.

\paragraph{Reinforcement Learning} ‌This task is implemented by training LLMs with reinforcement learning to identify and decline unknown questions. For instance, \citet{xu2024sayself} present the SaySelf training framework, which enhances the model's ability to express fine-grained confidence estimates by generating self-reflective rationales. Similarly, to enable models to express uncertainty in natural language when beyond their knowledge boundaries, \citet{cheng2024can} constructed the ``I don't know'' (Idk) dataset, and achieved this through alignment training. \citet{xu2024rejection} proposed the Reinforcement Learning with Knowledge Feedback (RLKF) method, helping models identify knowledge boundaries and reject out-of-scope questions, thereby implicitly quantifying uncertainty and improving reliability. To identify and address the systematic bias in reward models towards high confidence scores in RLHF training, \citet{leng2025taming} proposed calibrated reward methods such as PPO-M and PPO-C, thereby effectively reducing calibration error in LLMs.

\paragraph{Probing} Probing methods quantify uncertainty by analyzing internal model representations. For example, \citet{azaria2023internal} introduced the SAPLMA method, which predicts statement truth probabilities by analyzing hidden layer activations, providing a quantitative approach to assess LLMs' internal confidence. \citet{marks2024the} proposed mass-mean probing to identify and intervene in linear representations of factual truth within models, opening new avenues for quantifying confidence in factual claims.

\subsubsection{Training-Free Methods}
Training-free methods estimate uncertainty without additional optimization, leveraging existing model outputs or behaviors through two main approaches:

\paragraph{Prediction Probability} These methods estimate uncertainty by analyzing output probability distributions. A common method for measuring uncertainty is Predictive Entropy~\citep{settles2008an}. To enable comparison across different length outputs, \citet{malinin2021uncertainty} systematically applied length-normalization to various information-theoretic uncertainty measures, enhancing their practical utility. \citet{xiao2021on} further applied entropy values to the hallucination problem in conditional language generation tasks, revealing the key relationship between uncertainty and hallucination. One method is to prompt the extraction of uncertainty information through specially designed inputs. The P(True) method \citep{kadavath2022language}, for example, asks models to evaluate the probability of their answers being correct, with higher probabilities indicating lower uncertainty. This approach relies on models' self-assessment capabilities. Addressing the ``generative inequality'' problem in LLMs free-form output, where irrelevant tokens and sentences are over-valued when estimating uncertainty, \citet{daun2024shifting} proposed Shifting Attention to Relevance (SAR), a method that mitigates these biases by reassigning attention based on the jointly examined relevance of components during uncertainty quantification. To address the bias of Predictive Entropy when handling irrelevant information, \citet{wang2025word} proposed Word-Sequence Entropy (WSE). WSE enhances the accuracy and stability of uncertainty quantification by incorporating semantic relevance at both word and sequence levels, utilizing off-the-shelf LLMs and multi-sample generation. This approach also significantly improves LLMs' accuracy in medical question-answering applications.

\paragraph{Sampling \& Aggregation} These methods estimate uncertainty by generating and analyzing multiple outputs. For example,\citet{farquhar2024detecting} introduced Semantic Entropy to quantify uncertainty in generation tasks by considering semantic equivalence.To address the insufficiency and unreliability of entropy as a confidence metric, particularly under distribution shifts and biased scenarios, \citet{lee2024entropy} introduced the Pseudo-Label Probability Difference (PLPD) as a newly proposed confidence metric, allowing it to identify more reliable samples by leveraging information that entropy cannot capture.Meanwhile, \citet{xiong2024can} further systematically proposed a general black-box confidence elicitation framework, combining human-inspired prompting strategies, diverse sampling methods, and confidence aggregation mechanisms, which can significantly improve LLMs' performance in tasks like confidence calibration and failure prediction, even without internal access. \citet{aichberger2025improving} introduced the Semantically Diverse Language Generation (SDLG) method, which steers LLMs to generate semantically diverse yet likely alternative texts, significantly improving the estimation accuracy of predictive uncertainty. \citet{lyu2025calibrating} evaluated consistency-based metrics for post-hoc uncertainty quantification.

\subsection{Reranking Methods in RAG}
Initial retrieval in RAG still struggles with issues like irrelevant or suboptimal results, missing key documents, and data quality concerns—highlighting the importance of reranking optimization. These reranking methods fall into three categories: pointwise, pairwise, and listwise.

\subsubsection{Pointwise Methods}
Pointwise reranking ranks documents or items by independently scoring their relevance and sorting them. \citet{sachan2022improving} achieved unsupervised Top-20 accuracy improvements on BEIR benchmarks via pre-trained model-based probabilistic inference, which established zero-shot probabilistic ranking paradigms. \citet{bonifacio2022inpars} leveraged few-shot generation capabilities to build synthetic query-document pairs via monoT5. \citet{boytsov2024inparslight} extended this to industrial-grade InPars-Light through model compression and toolchain integration.

\subsubsection{Pairwise Methods}
Pairwise reranking predicts the relative order of document pairs to achieve overall document ranking. Its advantage lies in closely aligning with ranking’s core by focusing on relative document order, effectively minimizing inversions \citep{Joachims2002optimizing}. Recent progress addresses computational bottlenecks and probabilistic modeling. \citet{khattab2020colbert} improved efficiency via precomputed BERT document representations and late interaction. To address the order sensitivity of traditional methods, \citet{luo2024prp} introduced the PRP-Graph framework, integrating probabilistic relevance propagation, dynamic graph aggregation, and weighted PageRank to enhance ranking model stability. Similarly, \citet{li2024prd} proposed the PRD framework, introducing dynamic weighting mechanisms and multi-turn discussion strategies, which effectively mitigated the limitations of single-model evaluation and enhanced the adaptability and robustness of ranking systems. As generative retrieval (GR) develops, researchers began considering how to extend it to accommodate multi-graded relevance scenarios. In this context, \citet{tang2024generative} introduced a multi-graded constrained contrastive (MGCC) loss which incorporates grade penalty and constraint mechanisms. 
Concurrently, \citet{heinrich2025axiomatic} defined pairwise preferences based on the similarity between argumentative units and the query, aggregated these preference signals in the reranking phase, significantly enhancing argument retrieval effectiveness.
Addressing the effectiveness-efficiency trade-off optimization for retrieval using multiple prediction models, \citet{oosterhuis2025optimizing} introduced compound retrieval systems. These systems, via a selection policy, learn to acquire BM25 and LLM pointwise and pairwise predictions and aggregate them for final ranking construction, enhancing retrieval efficiency.

\subsubsection{Listwise Methods}
Listwise reranking optimizes the entire document lists to enhance retrieval effectiveness. To account for varying numbers of document pairs per query and more accurately capture ranking metrics, \citet{cao2007learning} proposed the listwise learning‑to‑rank approach. This approach uses document lists as instances in learning, contrasting with the pairwise approach which uses document pairs. By assigning different weights to training instances during optimization, ListMAP significantly improved document ranking effectiveness \citep{keshvari2022listmap}. To address issues like the discrepancy between LLM pre-training objectives and the ranking objective and the lack of direct passage ranking capability, \citet{sun2023is} proposed the instructional permutation generation approach to directly output passage rankings. They then leveraged the permutation outputs generated by ChatGPT through a permutation distillation technique to transfer its ranking capability to smaller, efficient specialized models, which significantly improved passage re-ranking performance. Furthermore, research has advanced in incorporating listwise principles into the GR paradigm. \citet{tang2024listwise} proposed ListGR, which views the retrieval task as a sequential learning process for generating a list of docids. Considering the sensitivity of LLMs to the order of input passages, \cite{zhang2024are} proposed a simple listwise sampling approach that effectively mitigates dependence on the position of ground truth evidence by shuffling the list of input passages multiple times and aggregating results. Efficiency was further addressed by \citet{chen2025tourrank}, who implemented tournament-style dynamic grouping in their TourRank system. Furthermore, \citet{suresh2025cornstack} curated CORNSTACK—a high‐quality dataset filtered by dual consistency—for contrastive training, yielding significant performance gains in embedding models and code rerankers on code retrieval tasks.
Subsequently, \citet{liu2024towards} proposes a label formulation integrating content-based relevance and engagement-based relevance, leveraging sigmoid transformations to enhance the ranking system's ability to differentiate content quality and improve controllability in ranking. \citet{chowdhury2025rankshap} developed RankSHAP, a Shapley value-based feature attribution method, to improve ranking model explainability. It significantly enhances feature attribution fidelity and explainability for information retrieval reranking models. Furthermore , \citet{ren2025self} proposed the SCaLR method, which introduces explicit list-view relevance scores for global ranking and uses self-generated point-view relevance assessments to calibrate the list-view relevance, enhancing global comparability on large candidate sets.

\section{Methodology}
\label{sec:methodology}

This section delineates the proposed LLM-Confidence Reranker algorithm. We first elaborate on the Maximum Semantic Cluster Proportion, a confidence modeling technique tailored for black-box large language models (LLMs), as detailed in \S\ref{sec:confidence}, with an emphasis on a semantic consistency approach. Subsequently, we outline the comprehensive procedure of the LCR algorithm in \S\ref{sec:algorithm}.

\subsection{Maximum Semantic Cluster Proportion}
\label{sec:confidence}

Confidence represents a language model's assurance in its generated output, whereas uncertainty reflects the lack of such assurance. To quantify confidence in black-box LLMs, we employ a method based on semantic consistency assessment. Drawing inspiration from Semantic Entropy (SE)~\citep{farquhar2024detecting}, which measures uncertainty via the entropy of semantically clustered outputs, we introduce the Maximum Semantic Cluster Proportion (MSCP) as a straightforward and efficient confidence metric for black-box LLMs. This metric leverages the inherent semantic comprehension of LLMs to evaluate agreement across multiple generated outputs, serving as a reliable indicator of relevance in retrieval-augmented generation (RAG) systems.

Let \(\phi\) denote the parameters of a pre-trained LLM, treated as a black-box system accessible solely through input-output interactions. For a given input \(x\) (e.g., a query or query-document pair), the LLM \(\phi\) produces outputs by sampling from its predictive distribution. To assess confidence, we apply multinomial sampling at temperature \(T = 1\) to generate \(K\) independent output sequences, denoted as \(\{t_k\}_{k=1}^K\), where each \(t_k\) is autoregressively sampled as follows:
\[
t_k \sim p_{\phi}(\cdot \mid x),
\]
with \(p_{\phi}(t \mid x)\) representing the probability distribution over sequences induced by the LLM \(\phi\).

These \(K\) sequences are clustered according to semantic equivalence, forming a partition of semantically similar outputs. Clustering utilizes the same LLM \(\phi\) to assess pairwise semantic relations, ensuring alignment within the model's semantic space without external dependencies. For each pair of sequences \((t_i, t_j)\) where \(i \neq j\), \(t_i \sim p_{\phi}(\cdot \mid x)\), and \(t_j \sim p_{\phi}(\cdot \mid x)\), we prompt the LLM \(\phi\) with a natural language inference (NLI)-style query to classify their semantic relation. This is formalized by a function \(f_{\phi}(t_i, t_j)\) that outputs one of: ``entailment'', ``contradiction'', or ``neutral'', indicating whether \(t_i\) semantically entails \(t_j\).

This pairwise assessment constructs an undirected similarity graph \(G = (V, E)\), where the vertices \(V = \{t_k\}_{k=1}^K\) comprise the sampled sequences, and an edge \((t_i, t_j) \in E\) exists if and only if bidirectional entailment holds:
\[
(t_i, t_j) \in E \iff f_{\phi}(t_i, t_j) = \text{``entailment''} \land f_{\phi}(t_j, t_i) = \text{``entailment''}.
\]

The connected subgraphs in graph \(G\) are defined as node subsets where a path---a sequence of edges---exists between every pair of nodes, and no additional nodes can be included while preserving this connectivity. Such maximal subgraphs form the connected components of \(G\), partitioning the graph into disjoint sets that represent distinct semantic equivalence classes. These classes aggregate responses sharing the same underlying meaning, with bidirectional entailment ensuring transitivity and symmetry within each component, thus propagating equivalence across connected nodes.

Direct computation of these connected components via a complete graph and standard algorithms (e.g., depth-first search or union-find) can be computationally demanding for large \(K\), requiring \(\mathcal{O}(K^2)\) pairwise entailment evaluations. To address this, we adopt a greedy clustering algorithm that approximates the graph structure without materializing all edges. The algorithm initializes an empty cluster set \(S = \emptyset\). For each sample \(t_i\) (processed sequentially from \(i=1\) to \(K\)):

\begin{itemize}
\item Assess whether \(t_i\) can be assigned to an existing cluster \(s \in S\). For each such cluster \(s\), select a representative response \(t^{(s)}\) from \(s\), such as the initial sample added or an embedding-derived centroid for efficiency.
\item If \(f_{\phi}(t_i, t^{(s)}) = \text{``entailment''} \land f_{\phi}(t^{(s)}, t_i) = \text{``entailment''}\), incorporate \(t_i\) into \(s\).
\item If \(t_i\) aligns with no existing cluster, initialize a new cluster \(\{t_i\}\) and append it to \(S\).
\end{itemize}

This greedy approach efficiently identifies connected components by iteratively merging samples based on entailment links, ensuring each resulting cluster corresponds to a maximal connected subgraph where all responses are semantically equivalent via transitive entailment paths. The algorithm reduces NLI evaluations (via LLM  \(\phi\)) to \(\mathcal{O}(K \cdot M)\), where \(M\) is the number of clusters (typically much smaller than \(K\)), rendering it suitable for practical deployment. The final semantic equivalence classes are \(\{s_1, s_2, \dots, s_M\}\), with \(M\) denoting the number of distinct meanings identified, and cluster sizes satisfying \(\sum_{m=1}^M |s_m| = K\), as each sample is assigned to exactly one class.

Following cluster formation, MSCP is formally defined as:
\begin{equation}
C(x) = \operatorname{MSCP}(x; \phi, K) = \max_{m=1,\dots,M} \frac{|s_m|}{K},
\label{eq:mscp}
\end{equation}
where \(|s_m|\) is the cardinality of the \(m\)-th cluster.

This metric quantifies confidence as the proportion of samples in the largest semantic cluster, with values near 1 indicating high semantic consistency and elevated confidence, and lower values reflecting greater dispersion across clusters, signifying uncertainty. Compared to SE, which computes entropy over all cluster proportions and may be sensitive to minor distributional variations, MSCP offers a more direct and interpretable measure by focusing solely on the dominant semantic mode. This simplicity facilitates faster computation and aligns well with reranking objectives in RAG systems, where prioritizing documents that elicit strongly consistent responses is essential.

By employing the same LLM for sampling and clustering, MSCP ensures internal consistency, reducing discrepancies from model mismatches and enhancing its utility in black-box scenarios. Empirical evaluations, detailed in subsequent sections, confirm that MSCP captures confidence signals correlated with document relevance, supporting the LCR algorithm's ranking improvements.

\begin{figure}[pos=htbp]
    \centering
    \includegraphics[width=\linewidth]{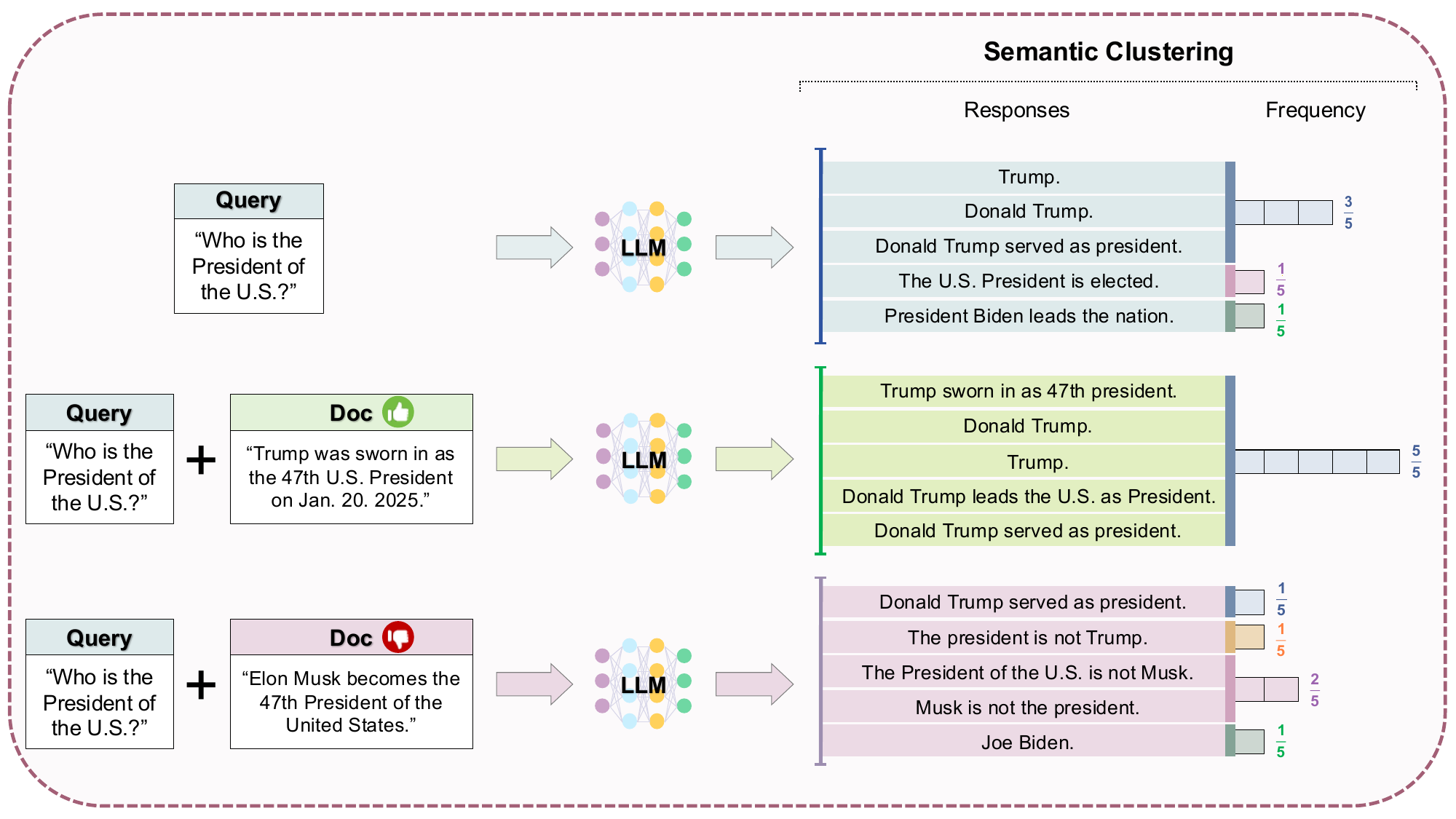}
    \caption{\textbf{Illustration of Maximum Semantic Cluster Proportion.} As illustrated in the figure, the question \( q \) is ``Who is the President of the U.S.?'' We instruct a specific LLM with parameters \( \phi \) to sample \( K \) times (where \( K = 5 \) in this example) at a high temperature (typically \( T = 1 \)). Through clustering, three semantic clusters are derived, with the largest semantic cluster being ``Trump'', which accounts for \( 3/5 \) of the responses, denoted as \( \text{MSCP}(q; \phi, K) = 3/5 \). When we provide both the question \( q \) and the document \( d_1 \), which states, ``Trump was sworn in as the 47th U.S. President on Jan. 20. 2025'', all five sampled responses consistently point to ``Trump'', resulting in \( \text{MSCP}(q, d_1; \phi, K) = 5/5 \). Compared to the scenario where only \( q \) is provided, the MSCP increases, indicating that \( d_1 \) is a helpful document for \( q \). In contrast, when we supply both the question \( q \) and the document \( d_2 \), which claims, ``Elon Musk becomes the 47th President of the United States'', the largest semantic cluster after sampling and clustering only constitutes \( 2/5 \),  denoted as \( \text{MSCP}(q, d_2; \phi, K) = 2/5 \). This represents a decrease in MSCP compared to when only \( q \) is provided, suggesting that \( d_2 \) is a harmful document for \( q \).}
    \label{fig:mscp}
\end{figure}

\subsection{LLM-Confidence Reranker Algorithm}
\label{sec:algorithm}

The LLM-Confidence Reranker (LCR) algorithm enhances document ranking via confidence interval binning and multi-level sorting. It defines a query confidence threshold \( T_{\text{query}} \), a high confidence threshold \( T_{\text{upper}} \), and a low confidence threshold \( T_{\text{lower}} \), with its core process detailed in Algorithm~\ref{alg:lcr} and Figure~\ref{fig:lcr}. In the algorithm, confidence for documents and queries is denoted by \(C(\cdot)\), prior relevance scores (e.g., from a retriever or reranker) by \(PrevScore(\cdot)\), and the document set by \( \mathcal{D} = \{d_i\}_{i=1}^n \).

\begin{algorithm}[!htbp]
\caption{The LLM-Confidence Reranker Algorithm}\label{alg:lcr}
\begin{algorithmic}[1]
\Function{BinnedConfidenceScore}{$q, d$}
    \If{$C(q,d) \geq T_{\text{upper}}$}
        \State \Return 1 \Comment{High confidence bin}
    \ElsIf{$C(q,d) \leq T_{\text{lower}}$}
        \State \Return -1 \Comment{Low confidence bin}
    \Else
        \State \Return 0 \Comment{Medium confidence bin}
    \EndIf
\EndFunction

\Function{LCR\_Sort}{$q, \mathcal{D}$}
    \State Generate joint features \( \{(q,d)\}_{d \in \mathcal{D}} \)
    \If{$C(q) < T_{\text{query}}$}
        \State $\mathcal{D}' \gets \text{StableSort}(\mathcal{D}, [\mathit{BinnedConfidenceScore}(q,d) \downarrow, \mathit{PrevScore}(q,d) \downarrow])$ 
    \Else
        \State $\mathcal{D}' \gets \text{Sort}(\mathcal{D}, \mathit{PrevScore}(q,d) \downarrow)$ 
    \EndIf
    \State \Return $\mathcal{D}'$
\EndFunction
\end{algorithmic}
\end{algorithm}

The algorithm operates through two integrated stages:
\begin{enumerate}
    \item \textbf{Confidence Assessment and Binning}: It first computes the query confidence \( C(q) \) and compares it to \( T_{\text{query}} \). Concurrently, it calculates joint confidence scores \( C(q,d) \) for each document \( d \) in \( \mathcal{D} \). Documents are then categorized into three bins using \( T_{\text{upper}} \) and \( T_{\text{lower}} \):
    \begin{itemize}
        \item \textbf{High confidence}: if \( C(q,d) \geq T_{\text{upper}} \)
        \item \textbf{Medium confidence}: if \( T_{\text{lower}} < C(q,d) < T_{\text{upper}} \)
        \item \textbf{Low confidence}: if \( C(q,d) \leq T_{\text{lower}} \)
    \end{itemize}
    \item \textbf{Ranking Strategy}: If \( C(q) < T_{\text{query}} \) (low confidence), documents are sorted first by confidence bins in descending order (high → medium → low), then by \( \mathit{PrevScore}(q,d) \) within each bin. If \( C(q) \geq T_{\text{query}} \) (high confidence), documents are sorted directly by \( \mathit{PrevScore}(q,d) \) in descending order.
\end{enumerate}

By discretizing confidence into three intervals, LCR reduces noise and sharpens distinctions between documents.

\section{Evaluation and Discussion}
\label{sec:experiments}

This section provides a thorough evaluation of the LLM-Confidence Reranker (LCR) algorithm, assessing its performance, robustness, and underlying mechanisms. We start with the experimental setup in \S\ref{sec:setup}, outlining the datasets, retrievers, rerankers, confidence quantification methods, and evaluation metrics employed. \S~\ref{sec:result} then presents the primary experimental results, comparing NDCG@5 scores across different configurations with BM25 as the initial retriever. The subsequent subsection analyzes the impact of different retrievers, incorporating Contriever to evaluate compatibility. Additional analyses examine the effects of the query threshold, sensitivity to document thresholds, variations in language models, and different confidence quantification approaches. Finally, we investigate the underlying mechanism of the LCR algorithm, as detailed in \S\ref{sec:mechanism}.

\subsection{Experiment Setup}
\label{sec:setup}
\subsubsection{Datasets}
We evaluate our approach on datasets from two prominent information retrieval benchmarks: BEIR~\citep{thakur2021beir} and TREC~\citep{voorhees2005trec}. BEIR (Benchmarking IR) is a heterogeneous zero-shot evaluation benchmark comprising datasets across diverse domains, designed to assess the generalization of retrieval models in out-of-domain settings. TREC (Text REtrieval Conference) is an annual series organized by NIST, providing standardized tasks and judgments for advancing retrieval research; we specifically use the Deep Learning tracks from 2019 and 2020, which focus on passage ranking with graded relevance annotations.
From BEIR, we select six datasets spanning various query types and relevance challenges, ideal for testing real-world retrieval performance in knowledge-intensive tasks:
\begin{itemize}
\item \textbf{NaturalQuestions} (NQ)~\citep{kwiatkowski2019natural}: An open-domain question-answering dataset with real user queries from Google search, annotated with answers from Wikipedia passages.
\item \textbf{DBpedia-Entity} (DBPE)~\citep{hasibi2017dbpedia}: An entity-centric retrieval dataset emphasizing entity linking and knowledge graph queries, with relevance based on DBpedia entity relationships.
\item \textbf{FEVER} (FEVE)~\citep{thorne2018fever}: A fact verification dataset where queries are claims verified against Wikipedia documents; relevance labels denote support, refutation, or neutrality.
\item \textbf{SciDocs} (SCID)~\citep{cohan2020specter}: A citation prediction dataset using scientific paper abstracts as queries, with relevance determined by citation links.
\item \textbf{Touché} (TOUC)~\citep{bondarenko2020overview}: An argument retrieval dataset featuring debate topics as queries, with relevance tied to supporting or opposing arguments in web documents.
\item \textbf{NFCorpus} (NFCO)~\citep{boteva2016full}: A biomedical information retrieval dataset with medical queries and scientific articles, where expert-assigned labels indicate document relevance.
\end{itemize}

From TREC, we incorporate two datasets:

\begin{itemize}
\item \textbf{TREC DL19} (DL19)~\citep{craswell2020overview}: Passage ranking task from TREC 2019 Deep Learning track, with queries and graded relevance judgments on a large corpus.
\item \textbf{TREC DL20} (DL20)~\citep{craswell2021overview}: Similar to DL19 but from the 2020 track, emphasizing neural retrieval methods with diverse query-document pairs.
\end{itemize}

We utilize the test sets and relevance annotations provided by these benchmarks.

\subsubsection{Retrievers and Rerankers}
We employ BM25~\citep{jones2000probabilistic} as a classic sparse retriever based on term frequency and inverse document frequency, and Contriever~\citep{izacard2021contriever} as a dense retriever enhanced by contrastive learning. These retrievers are selected to evaluate LCR's compatibility across both sparse and dense retrieval paradigms, ensuring a comprehensive assessment of its plug-and-play nature in diverse RAG settings.

\begin{table}[htbp]
\centering
\caption{\textbf{Comparison of Reranking Methods.}}
\label{tab:rerankers}
\resizebox{\textwidth}{!}{
\begin{tabular}{lcccccc}
\toprule
Reranker &  Requires Internal Param. & Base Model & Architecture Type & Fine-tuned \\
\midrule
LCR (ours) & No & Qwen7B & Decoder-Only & No \\
QLM & Yes & Qwen7B & Decoder-Only & No \\
RankGPT & No & Qwen7B & Decoder-Only & No \\
YesNo & Yes & Qwen7B & Decoder-Only & No \\
Cross-Encoder & No & BERT-base & Encoder-Only & Yes \\
ColBERT & No & BERT-base & Encoder-Only & Yes \\
RankT5 & No & T5-base & Encoder-Decoder & Yes \\
\bottomrule
\end{tabular}
}
\end{table}

For rerankers, we select a diverse set including pre-trained LLM-based and fine-tuned Transformer-based methods to benchmark LCR's performance, as summarized in Table~\ref{tab:rerankers}. These rerankers are:
\begin{itemize}
	\item \textbf{QLM} \citep{sachan2022improving}: A point-wise reranking approach that prompts pre-trained language models to generate a relevant query for each candidate document, then ranks based on the likelihood of the actual query, without fine-tuning.
	\item \textbf{RankGPT} \citep{sun2023is}: A generative list-wise reranking method that prompts pre-trained language models to output a ranked list of document labels by relevance, using sliding windows, without fine-tuning.
    \item \textbf{YesNo} \citep{qin2024large}: A point-wise reranking technique that prompts pre-trained language models to generate ``yes'' or ``no'' indicating document relevance to the query, then ranks based on normalized ``yes'' likelihood, without fine-tuning.
    \item \textbf{Cross-Encoder} \citep{reimers2019sentence}: A cross-encoder method that computes relevance scores from joint query-document representations.
    \item \textbf{ColBERT} \citep{khattab2020colbert}: A context embedding-based reranking model that enhances precision through token-level interactions.
    \item \textbf{RankT5} \citep{zhuang2023rankt5}: A Transformer-based sequence-to-sequence model designed for reranking tasks.
\end{itemize}

These rerankers are chosen to compare LCR---a training-free, confidence-based method---against both similar pre-trained LLM-based approaches (e.g., QLM, RankGPT, YesNo) that underutilize semantic understanding, and fine-tuned Transformer models that represent optimized, domain-specific baselines, thereby highlighting LCR's advantages in generalization and efficiency.

For each query, the retriever fetches top-10 documents, reranked by the specified method, including LCR. For LLM-based rerankers, we use Qwen2.5-7B-Instruct (Qwen7B)~\citep{yang2024qwen2}. We employ MSCP, as detailed in Equation~\ref{eq:mscp}, for confidence quantification.

\subsubsection{Confidence Quantification Setup}
Multinomial sampling ($T=1$) generates $K=10$ samples per query-document pair.
Sampling prompts:

\begin{itemize}
    \item Without document:
        \begin{grayquote}
        Answer the following question as briefly as possible.\\
        Question: \{query\}\\
        Answer:
        \end{grayquote}
    \item With document:
        \begin{grayquote}
        Answer the following question as briefly as possible.\\
        Context: \{document\}\\
        Question: \{query\}\\
        Answer:
        \end{grayquote}
\end{itemize}

Clustering evaluates semantic relationships:
\begin{grayquote}
We are evaluating answers to the question ``\{query\}''\\
Possible Answer 1: \{answer1\}\\
Possible Answer 2: \{answer2\}\\
Does Possible Answer 1 semantically entail Possible Answer 2? Respond with only one of the following words: entailment, contradiction, or neutral. Do not provide any additional explanation.\\
Response:
\end{grayquote}

\subsubsection{Evaluation Metrics}
We use Normalized Discounted Cumulative Gain (NDCG)~\citep{jarvelin2002cumulated} at $k=5$ as the primary metric, prioritizing relevant documents early for RAG systems.

NDCG@k evaluates ranking quality by accounting for document relevance and position. It is computed as Discounted Cumulative Gain (DCG) up to position $k$, normalized by the Ideal DCG (IDCG) for the optimal ranking:
\[
\text{NDCG@k} = \frac{\text{DCG@k}}{\text{IDCG@k}}, \quad \text{DCG@k} = \sum_{i=1}^{k} \frac{\text{rel}_i}{\log_2(i+1)},
\]
where $\text{rel}_i$ denotes the relevance of the document at position $i$. Higher NDCG@k values indicate superior performance, prioritizing relevant documents early.

\subsection{Experiment Results}
\label{sec:result}

Table~\ref{tab:performance-bm25} presents the NDCG@5 performance comparison when using BM25 as the initial retriever, combined with various rerankers, and further enhanced by LCR across datasets. As a plug-and-play algorithm based on LLM confidence signals, LCR ensures a safe performance lower bound: when the query confidence threshold \(T_{\text{query}}=0\), the algorithm reverts to the original ranking, maintaining performance at least at the baseline level in the worst case. This design provides high robustness for LCR in practical deployment, avoiding potential risks. The values in the table represent the best performance after tuning the three hyperparameters (\(T_{\text{query}}\), \(T_{\text{upper}}\), \(T_{\text{lower}}\)), aimed at demonstrating LCR's upper potential; subsequent sensitivity analysis in Sections~\ref{sec:impact-of-query-threshold} and~\ref{sec:sensitivity-analysis-of-document-thresholds} will further validate its generalization ability.

The improvements from LCR exhibit high consistency: across all reranker and dataset combinations, adding LCR results in NDCG@5 gains or ties, with no instances of decline. This arises from LCR's core mechanism: through confidence binning, it prioritizes high-confidence documents at the top, while retaining the original ranking when query confidence is high to avoid unnecessary interference. The averages across BEIR and TREC datasets indicate relative gains of up to 20.6\% (e.g., YesNo on BEIR: 0.1737 to 0.2095) from LCR, with no declines. This stable enhancement confirms the hypothesis that LLM confidence correlates positively with document relevance (detailed in Section~\ref{sec:mechanism}).

From the perspective of dataset heterogeneity, LCR's gains are particularly notable on factual or argumentative datasets (e.g., NaturalQuestions, Touché, FEVER), such as RankGPT on NaturalQuestions improving from 0.0489 to 0.0587 (a 20\% relative increase). This reflects that such queries heavily rely on semantic consistency from external documents, where LCR's confidence signals effectively amplify relevance differences. On domain-specific datasets with already high baseline performance (e.g., FEVER + fine-tuned Transformer-based rerankers), LCR maintains stability without degradation, demonstrating its adaptability. For TREC datasets (DL19 and DL20), LCR's improvements are equally reliable, for instance, from 0.5172 to 0.5271 (a 1.9\% relative increase) when applying LCR directly without any reranker. The document diversity and long-tail query characteristics of TREC datasets further highlight LCR's generalization advantage in handling complex retrieval scenarios.

Breaking down by reranker type reveals LCR's strong compatibility. First, in the pure BM25 Retriever-Only scenario, LCR yields average gains of 3.0\% (BEIR) and 1.9\% (TREC), with slight optimizations across all datasets, proving it can independently improve sparse retrieval results without relying on additional rerankers. This makes it especially suitable for resource-constrained settings. Second, for pre-trained LLM-based rerankers (YesNo, QLM, RankGPT, using 7B-parameter models), baseline performance is often below or close to Retriever-Only (BEIR average 0.1737--0.2391), reflecting limitations in ranking ability for small-scale models; however, LCR significantly reverses this disadvantage, such as a 20.6\% average gain on BEIR for YesNo. This suggests that even with limited model scale, the core semantic understanding can capture document ``helpfulness'' through confidence signals, thereby amplifying ranking effects. Finally, for fine-tuned Transformer-based rerankers (ColBERT, Cross-Encoder, RankT5), LCR's gains are smaller (about 1\%), but highly consistent. This verifies LCR's good compatibility with domain-specific training, allowing further refinement of document relevance on high-performance baselines.

In summary, as a training-free and lightweight (7B-parameter) algorithm, LCR reliably enhances document relevance in RAG systems, with an average gain of about 3\% across methods and datasets, thereby reducing hallucination risks. Its lower-bound guarantee and plug-and-play nature make it particularly suitable for production-level knowledge-intensive tasks.




\begin{table}[htbp]
\centering
\caption{\textbf{Comparative Performance of BM25 with Various Rerankers and LCR Enhancement.} NDCG@5 scores for the BM25 retriever combined with various rerankers, with and without the LCR (LLM-Confidence Reranker) enhancement, across BEIR datasets (NQ: NaturalQuestions, TOUC: Touché, SCID: SciDocs, NFCO: NFCorpus, DBPE:
DBpedia-Entity, FEVE: FEVER, AVG: average across BEIR datasets) and TREC datasets (DL19, DL20, AVG: average across TREC datasets). Bold values indicate the best performance within each reranker group, and underlined values indicate the best performance between the same reranker with and without LCR.}
\label{tab:performance-bm25}
\resizebox{\textwidth}{!}{
\begin{tabular}{lcccccccccc}
\toprule
\multicolumn{1}{c}{} & \multicolumn{7}{c}{\textbf{BEIR}} & \multicolumn{3}{c}{\textbf{TREC}} \\
\cmidrule(lr){2-8} \cmidrule(lr){9-11}
\textbf{Method} & \textbf{NQ} & \textbf{TOUC} & \textbf{SCID} & \textbf{NFCO} & \textbf{DBPE} & \textbf{FEVE} & \textbf{AVG} & \textbf{DL19} & \textbf{DL20} & \textbf{AVG} \\
\midrule
\multicolumn{11}{l}{\textbf{Retriever-Only}} \\ \midrule
- & .0488 & .4911 & .1144 & .3640 & .1059 & .3111 & .2392 & .5278 & .5067 & .5172 \\
\hspace{1em}+LCR & \textbf{\underline{.0585}} & \textbf{\underline{.5015}} & \textbf{\underline{.1149}} & \textbf{\underline{.3659}} & \textbf{\underline{.1078}} & \textbf{\underline{.3294}} & \textbf{\underline{.2463}} & \textbf{\underline{.5431}} & \textbf{\underline{.5112}} & \textbf{\underline{.5271}} \\
\midrule
\multicolumn{11}{l}{\textbf{Pre-trained LLM Based Rerankers}} \\ \midrule
YesNo & .0321 & .4388 & .0685 & .2941 & .0777 & .1310 & .1737 & .5925 & .5485 & .5705 \\
\hspace{1em}+LCR & \underline{.0516} & \underline{.4600} & \underline{.0872} & \underline{.3087} & \underline{.0912} & \underline{.2581} & \underline{.2095} & \textbf{\underline{.5974}} & \textbf{\underline{.5515}} & \textbf{\underline{.5744}} \\
\midrule
QLM & .0610 & .4090 & .1215 & .3415 & .1003 & .3595 & .2321 & .5341 & .4815 & .5078 \\
\hspace{1em}+LCR & \textbf{\underline{.0630}} & \underline{.4456} & \textbf{\underline{.1228}} & \underline{.3427} & \underline{.1037} & \textbf{\underline{.3608}} & \underline{.2397} & \underline{.5563} & \underline{.4898} & \underline{.5230} \\
\midrule
RankGPT & .0489 & .4914 & .1144 & .3630 & .1058 & .3111 & .2391 & .5278 & .5067 & .5172 \\
\hspace{1em}+LCR & \underline{.0587} & \textbf{\underline{.5018}} & \underline{.1149} & \textbf{\underline{.3649}} & \textbf{\underline{.1078}} & \underline{.3295} & \textbf{\underline{.2463}} & \underline{.5431} & \underline{.5103} & \underline{.5267} \\
\midrule
\multicolumn{11}{l}{\textbf{Fine-tuned Transformer Based Rerankers}} \\ \midrule
ColBERT & .0791 & .4778 & .1204 & .3832 & .1410 & \underline{.4745} & .2793 & \underline{.6377} & \underline{.6074} & \underline{.6226} \\
\hspace{1em}+LCR & \underline{.0805} & \underline{.4936} & \underline{.1205} & \underline{.3839} & \underline{.1413} & \underline{.4745} & \underline{.2824} & \underline{.6377} & \underline{.6074} & \underline{.6226} \\
\midrule
Cross-Encoder & .0812 & .4664 & \underline{.1266} & .3921 & .1465 & \underline{.4779} & .2818 & \underline{.6508} & .5983 & .6245 \\
\hspace{1em}+LCR & \underline{.0817} & \underline{.4972} & \underline{.1266} & \underline{.3936} & \underline{.1467} & \underline{.4779} & \underline{.2873} & \underline{.6508} & \underline{.5986} & \underline{.6247} \\
\midrule
RankT5 & .0831 & .5165 & .1376 & .4041 & .1479 & \textbf{\underline{.4864}} & .2959 & \textbf{\underline{.6548}} & .6065 & .6306 \\
\hspace{1em}+LCR & \textbf{\underline{.0839}} & \textbf{\underline{.5322}} & \textbf{\underline{.1378}} & \textbf{\underline{.4049}} & \textbf{\underline{.1482}} & \textbf{\underline{.4864}} & \textbf{\underline{.2989}} & \textbf{\underline{.6548}} & \textbf{\underline{.6078}} & \textbf{\underline{.6313}} \\
\bottomrule
\end{tabular}
} 
\end{table}

\subsection{Impact of Different Retrievers}
To assess LCR's compatibility with diverse retrievers, we employ Contriever~\citep{izacard2021contriever} as an additional initial retriever, which leverages contrastive learning to improve dense retrieval. Table~\ref{tab:performance-contriever} presents the NDCG@5 scores for Contriever combined with various rerankers, both with and without LCR enhancement (where ``+LCR w/o QT'' denotes LCR applied without the query threshold, and ``+LCR'' includes the query threshold; the impact of QT is discussed in \S\ref{sec:impact-of-query-threshold}).

The results indicate that LCR delivers consistent improvements over the Contriever baseline: across all reranker-dataset combinations, LCR achieves NDCG@5 gains or maintains equivalence, with no degradation. LCR achieves relative gains of up to 32.4\% on BEIR and TREC datasets, e.g., from 0.2281 to 0.3021 in the YesNo group (BEIR). Relative to BM25, Contriever exhibits a stronger baseline (BEIR average: 0.3939 vs. 0.2392); nevertheless, LCR's advantages are more pronounced for weaker rerankers such as YesNo, underscoring its capacity to amplify semantic signals and mitigate biases in dense retrieval.

Analysis across datasets reveals amplified gains on factual-oriented tasks (e.g., NQ, FEVER), as evidenced by RankGPT on NQ (1.1\% relative gain), which highlights the efficacy of confidence binning in elevating document relevance. On datasets with high baselines, such as FEVER, LCR preserves stability without excessive intervention. In conclusion, LCR performs robustly with both BM25 and Contriever, yielding an average 3.6\% gain and confirming its plug-and-play adaptability across sparse and dense retrieval paradigms.







\begin{table}[htbp]
\centering
\caption{\textbf{Comparative Performance of Contriever with Various Rerankers and LCR Enhancement.} NDCG@5 scores for the Contriever retriever combined with various rerankers, with and without LCR enhancement, across datasets (as defined in Table~\ref{tab:performance-bm25}). Here, ``+LCR w/o QT'' denotes LCR without the query threshold, and ``+LCR'' includes it. Bold and underlined values follow the conventions in Table~\ref{tab:performance-bm25}, with AVG denoting the average across datasets.}
\label{tab:performance-contriever}
\resizebox{\textwidth}{!}{
\begin{tabular}{lcccccccccc}
\toprule
\multicolumn{1}{c}{} & \multicolumn{7}{c}{\textbf{BEIR}} & \multicolumn{3}{c}{\textbf{TREC}} \\
\cmidrule(lr){2-8} \cmidrule(lr){9-11}
\textbf{Method} & \textbf{NQ} & \textbf{TOUC} & \textbf{SCID} & \textbf{NFCO} & \textbf{DBPE} & \textbf{FEVE} & \textbf{AVG} & \textbf{DL19} & \textbf{DL20} & \textbf{AVG} \\
\midrule
\multicolumn{11}{l}{\textbf{Retriever-Only}} \\ \midrule
- & .4554 & .2612 & .1180 & .3580 & .4285 & .7425 & .3939 & .6858 & .4575 & .5717 \\
\hspace{1em}+LCR w/o QT & \textbf{\underline{.4607}} & .2675 & .1185 & .3580 & .4294 & \textbf{\underline{.7431}} & .3962 & .6974 & \textbf{\underline{.4710}} & .5842 \\
\hspace{1em}+LCR & \textbf{\underline{.4607}} & \textbf{\underline{.2804}} & \textbf{\underline{.1195}} & \textbf{\underline{.3589}} & \textbf{\underline{.4364}} & \textbf{\underline{.7431}} & \textbf{\underline{.3998}} & \textbf{\underline{.6979}} & \textbf{\underline{.4710}} & \textbf{\underline{.5845}} \\
\midrule
\multicolumn{11}{l}{\textbf{Pre-trained LLM Based Rerankers}} \\ \midrule
YesNo & .2312 & .2476 & .0778 & .2645 & .2928 & .2546 & .2281 & .6990 & .5285 & .6137 \\
\hspace{1em}+LCR w/o QT & \underline{.3515} & .2577 & \underline{.0937} & \underline{.2882} & \underline{.3512} & \underline{.4689} & .3019 & .7088 & .5317 & .6202 \\
\hspace{1em}+LCR & \underline{.3515} & \underline{.2590} & \underline{.0937} & \underline{.2882} & \underline{.3512} & \underline{.4689} & \underline{.3021} & \textbf{\underline{.7089}} & \textbf{\underline{.5373}} & \textbf{\underline{.6231}} \\
\midrule
QLM & .3486 & .2269 & .1218 & .3143 & .3314 & .5569 & .3166 & .6577 & .4460 & .5519 \\
\hspace{1em}+LCR w/o QT & \underline{.3985} & \underline{.2599} & \textbf{\underline{.1224}} & .3143 & \underline{.3708} & \underline{.5871} & .3422 & .6821 & \underline{.4783} & .5802 \\
\hspace{1em}+LCR & \underline{.3985} & \underline{.2599} & \textbf{\underline{.1224}} & \underline{.3153} & \underline{.3708} & \underline{.5871} & \underline{.3423} & \underline{.6848} & \underline{.4783} & \underline{.5815} \\
\midrule
RankGPT & .4563 & .2612 & .1180 & .3569 & .4286 & .7427 & .3939 & .6858 & .4575 & .5717 \\
\hspace{1em}+LCR w/o QT & \textbf{\underline{.4613}} & .2675 & .1185 & .3569 & .4295 & \textbf{\underline{.7432}} & .3962 & .6974 & \underline{.4710} & .5842 \\
\hspace{1em}+LCR & \textbf{\underline{.4613}} & \textbf{\underline{.2804}} & \underline{.1195} & \textbf{\underline{.3579}} & \textbf{\underline{.4366}} & \textbf{\underline{.7432}} & \textbf{\underline{.3998}} & \underline{.6979} & \underline{.4710} & \underline{.5845} \\
\midrule
\multicolumn{11}{l}{\textbf{Fine-tuned Transformer Based Rerankers}} \\ \midrule
ColBERT & .4778 & .3570 & .1247 & .3574 & \underline{.4376} & .7617 & .4194 & .7193 & .6030 & .6612 \\
\hspace{1em}+LCR w/o QT & .4775 & .3570 & .1249 & .3574 & .4375 & \underline{.7619} & .4193 & .7278 & .6030 & .6654 \\
\hspace{1em}+LCR & \underline{.4808} & \textbf{\underline{.3651}} & \underline{.1254} & \underline{.3579} & \underline{.4376} & \underline{.7619} & \underline{.4214} & \underline{.7347} & \underline{.6039} & \underline{.6693} \\
\midrule
Cross-Encoder & .4870 & .3172 & .1306 & .3659 & .4595 & \underline{.7846} & .4241 & .7296 & .6089 & .6692 \\
\hspace{1em}+LCR w/o QT & .4867 & .3172 & .1308 & .3659 & \underline{.4599} & \underline{.7846} & .4242 & .7373 & .6101 & .6737 \\
\hspace{1em}+LCR & \underline{.4893} & \underline{.3244} & \underline{.1312} & \underline{.3665} & \underline{.4599} & \underline{.7846} & \underline{.4260} & \textbf{\underline{.7445}} & \underline{.6129} & \textbf{\underline{.6787}} \\
\midrule
RankT5 & .5097 & .3592 & \textbf{\underline{.1391}} & .3769 & .4679 & \textbf{\underline{.8156}} & .4447 & .7166 & .6103 & .6634 \\
\hspace{1em}+LCR w/o QT & .5096 & .3578 & .1389 & .3757 & \textbf{\underline{.4682}} & \textbf{\underline{.8156}} & .4443 & .7286 & .6115 & .6700 \\
\hspace{1em}+LCR & \textbf{\underline{.5106}} & \underline{.3615} & \textbf{\underline{.1391}} & \textbf{\underline{.3782}} & \textbf{\underline{.4682}} & \textbf{\underline{.8156}} & \textbf{\underline{.4455}} & \underline{.7336} & \textbf{\underline{.6130}} & \underline{.6733} \\
\bottomrule
\end{tabular}
} 
\end{table}

\subsection{Impact of Query Threshold}
\label{sec:impact-of-query-threshold}

To assess the impact of the query threshold (QT) on LCR performance, we first examine the results in Table~\ref{tab:performance-contriever}, which contrasts ``+LCR w/o QT'' and ``+LCR'' (with QT) across rerankers using Contriever as the retriever. On average, ``+LCR'' achieves higher scores on both BEIR and TREC (e.g., 0.3998 vs. 0.3962 for Retriever-Only on BEIR, 0.5845 vs. 0.5842 on TREC). These patterns indicate that QT further enhances LCR's ranking performance across various rerankers. Specifically, when the LLM exhibits low confidence in a query, indicating limited internal knowledge of the answer, it can more effectively distinguish helpful documents, leading to stronger confidence signals and greater gains from LCR. Conversely, high query confidence suggests the LLM already knows the answer, resulting in weaker signals for document helpfulness differentiation. Thus, QT enables selective application of LCR to queries where it provides the most benefit.

\begin{figure}[pos=htbp]
\begin{tabular}{@{}c@{}c@{}}
\begin{subfigure}{0.5\linewidth}
\centering
\includegraphics[width=\linewidth]{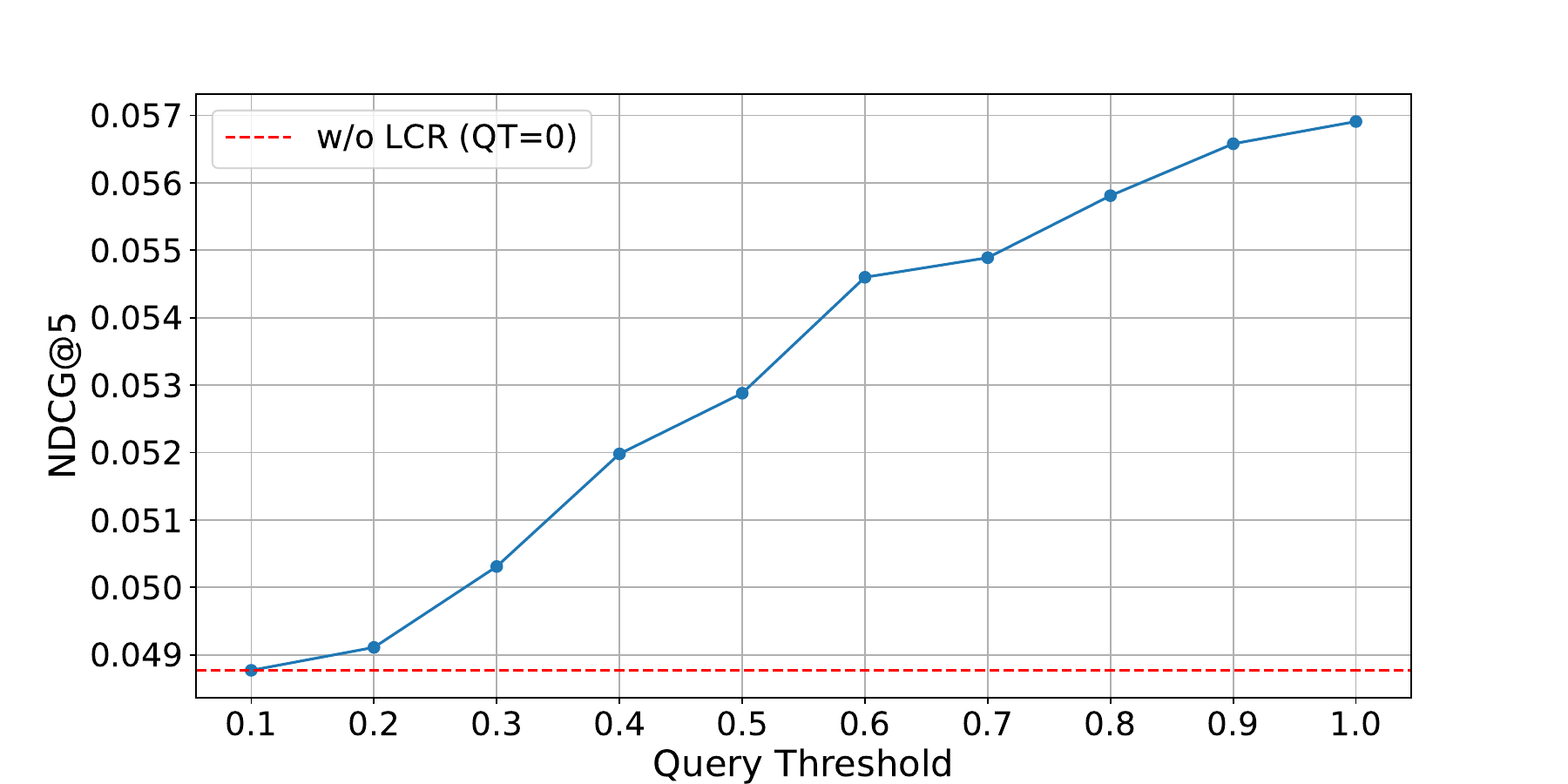}
\caption{Retriever-Only}
\label{fig:qt-nq-bm25-none}
\end{subfigure} &
\begin{subfigure}{0.5\linewidth}
\centering
\includegraphics[width=\linewidth]{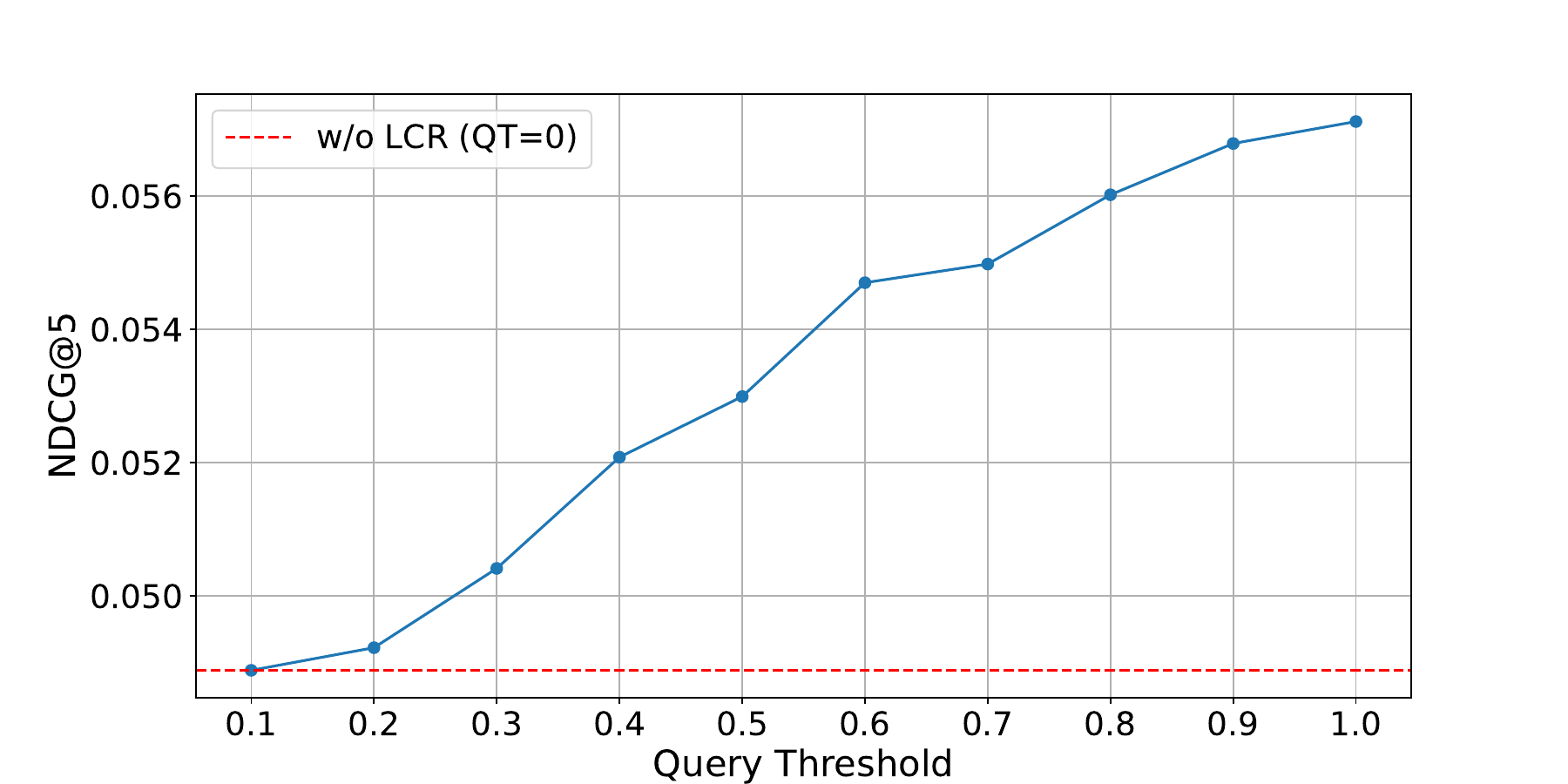}
\caption{RankGPT}
\label{fig:qt-nq-bm25-rankgpt}
\end{subfigure} \\
\begin{subfigure}{0.5\linewidth}
\centering
\includegraphics[width=\linewidth]{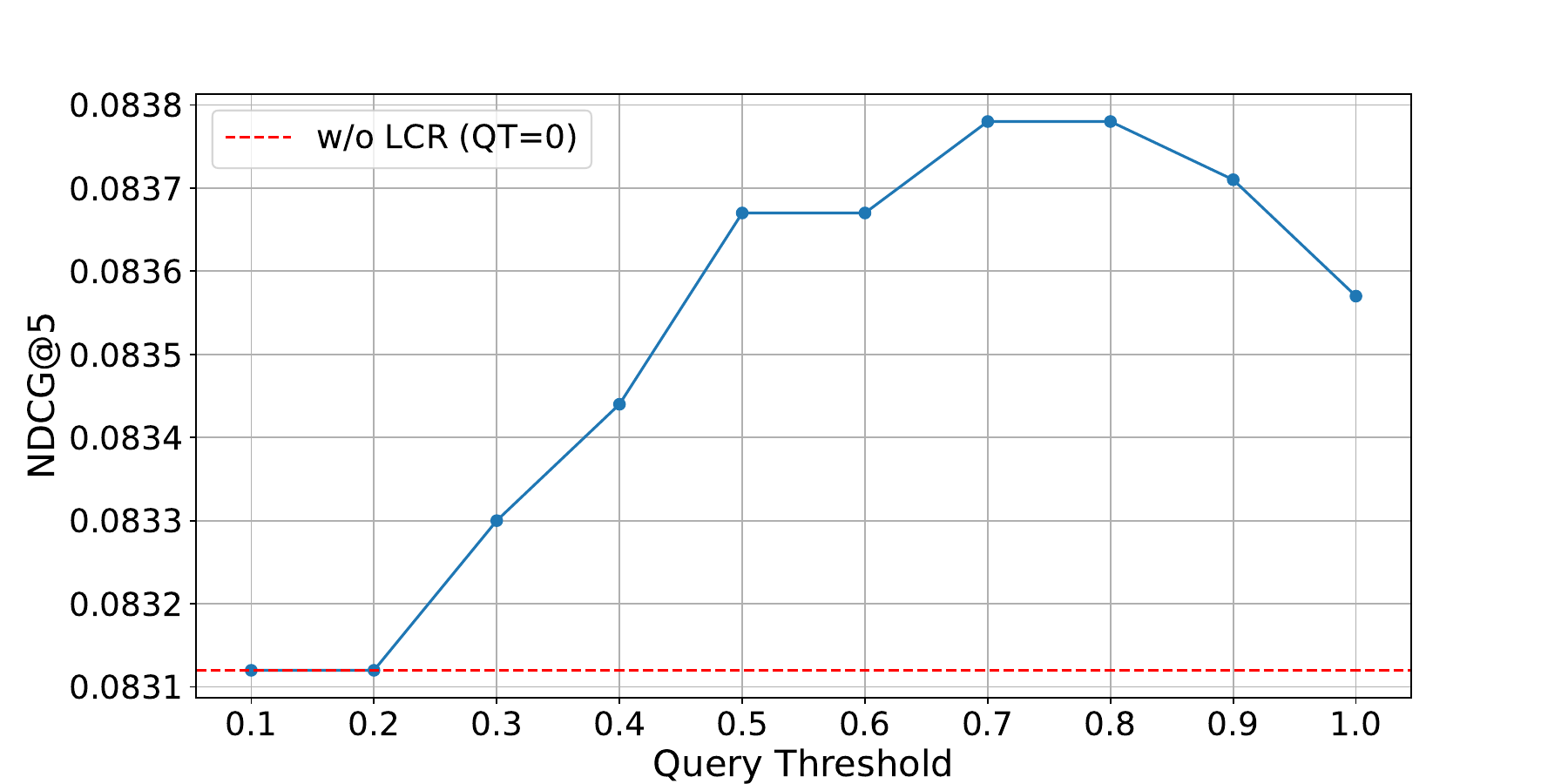}
\caption{RankT5}
\label{fig:qt-nq-bm25-rankt5}
\end{subfigure} &
\begin{subfigure}{0.5\linewidth}
\centering
\includegraphics[width=\linewidth]{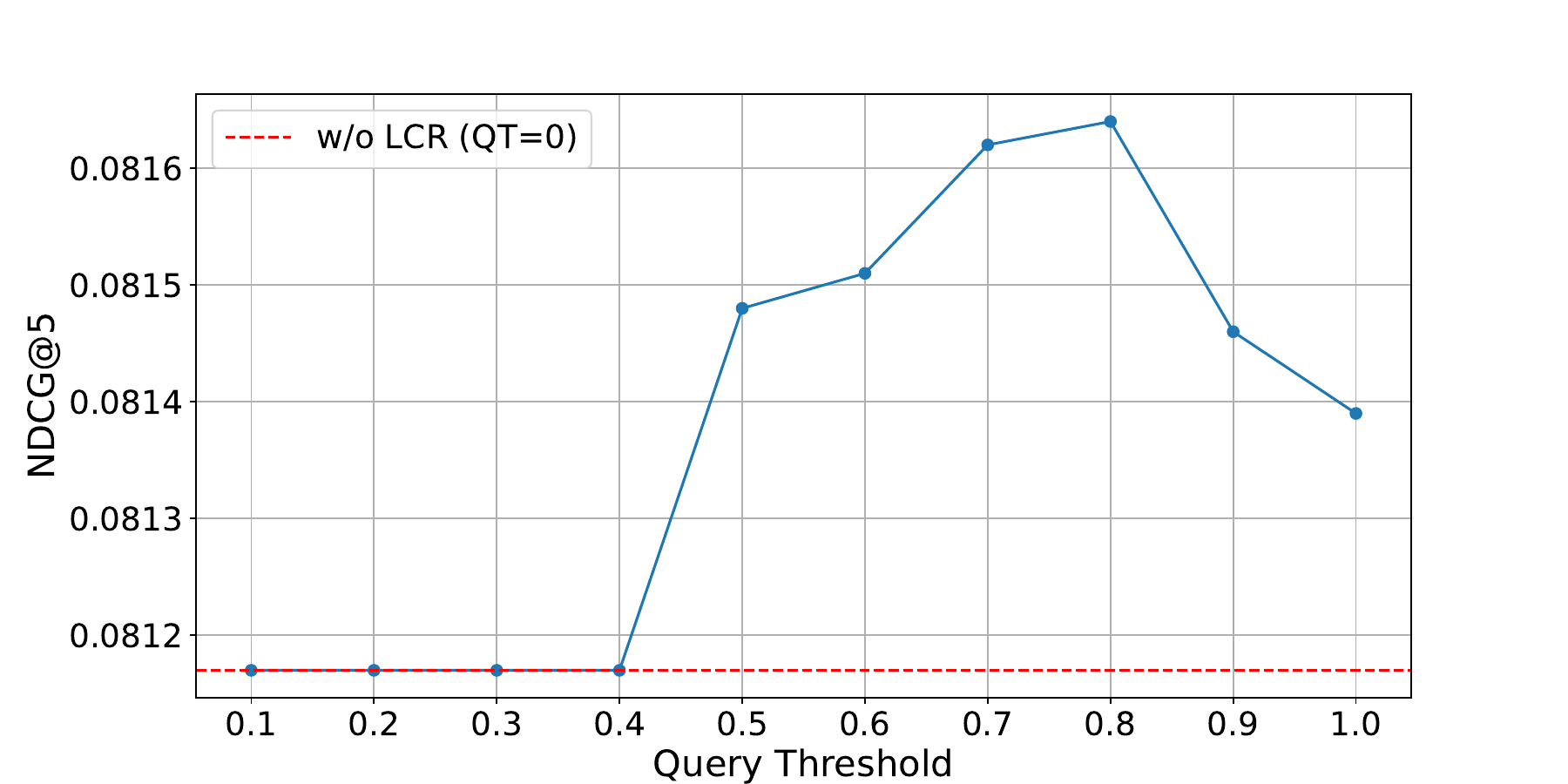}
\caption{Cross-Encoder}
\label{fig:qt-nq-bm25-crossencoder}
\end{subfigure}
\end{tabular}
\caption{\textbf{Query Threshold Impact on NDCG@5 for NaturalQuestions with BM25 Retriever.} These plots illustrate the impact of varying query threshold (QT) values on performance for different rerankers. The red dashed line indicates the baseline without LCR (QT=0).}
\label{fig:qt-nq-bm25}
\end{figure}

We further evaluate QT's influence on the NaturalQuestions dataset using BM25 as the retriever, with rerankers including Retriever-Only, RankGPT, Cross-Encoder, and RankT5. Figure~\ref{fig:qt-nq-bm25} shows NDCG@5 scores for QT values from 0.1 to 1.0, with QT=0 as the baseline (no LCR). For Retriever-Only and RankGPT, which rely on pre-trained LLMs and have weaker baselines, NDCG@5 rises progressively with QT, peaking at QT=1.0, as higher QT applies LCR to more queries for amplified gains. In contrast, RankT5 and Cross-Encoder, as fine-tuned rerankers with stronger baselines, exhibit an initial increase followed by a slight decline, requiring only slight QT adjustments. This suggests that higher QT benefits simpler rerankers by broadening confidence utilization, while advanced rerankers need minimal tuning to avoid unnecessary interference. Overall, optimal QT selection can significantly improve ranking in RAG systems.

\subsection{Sensitivity Analysis of Document Thresholds}
\label{sec:sensitivity-analysis-of-document-thresholds}

\begin{figure}[pos=htbp]
    \centering
    \includegraphics[width=1.0\linewidth]{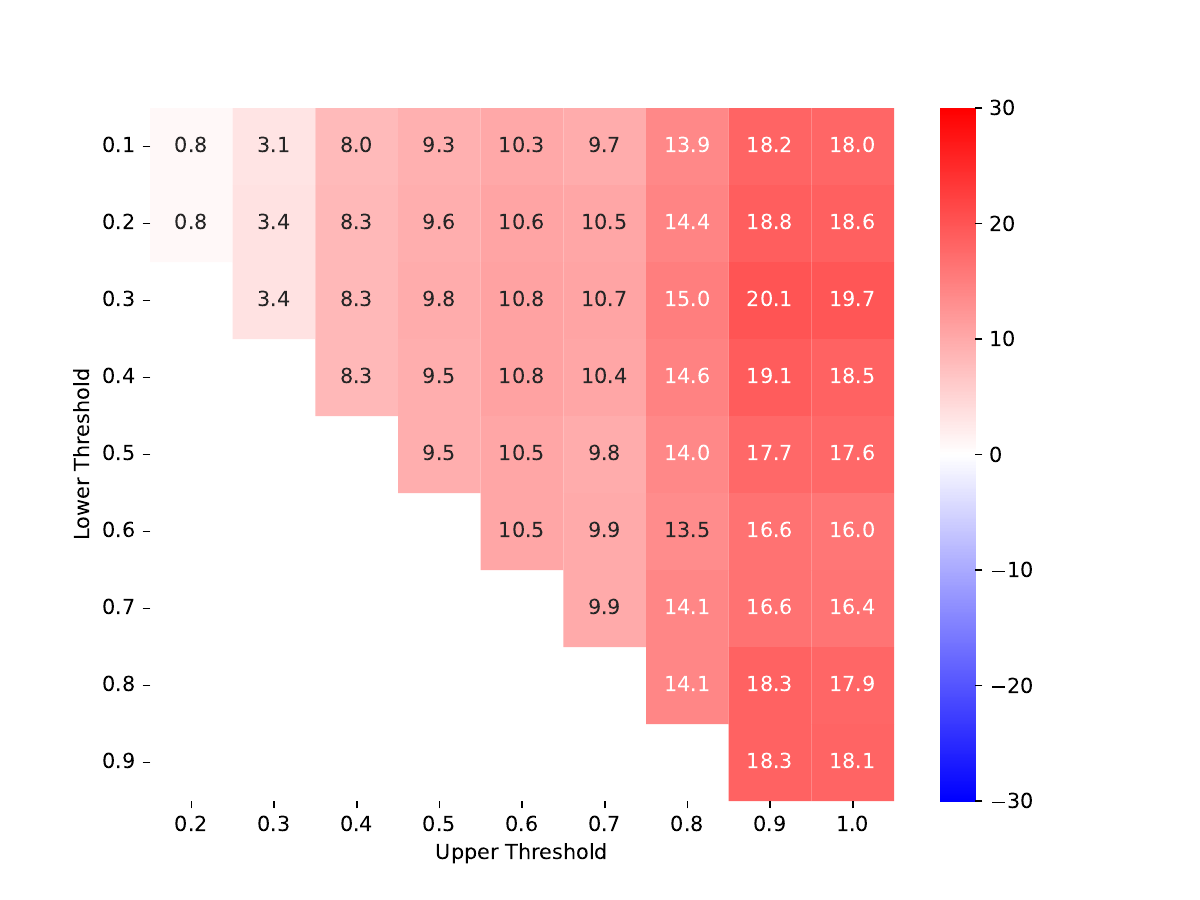}
    \caption{\textbf{Heatmap of NDCG@5 Improvement Percentages for Document Thresholds.} This heatmap shows the percentage differences in NDCG@5 scores for various combinations of the lower threshold (LT, y-axis) and upper threshold (UT, x-axis) in the LCR algorithm, applied to the NaturalQuestions dataset with BM25 as the initial retriever and RankGPT as the reranker, without the query threshold. Red shades indicate positive improvements relative to the baseline, blue shades signify negative changes, and white denotes no change. Color intensities correspond to the magnitude of variations, revealing optimal threshold settings for improved ranking performance.}
    \label{fig:param-dt}
\end{figure}

To assess the sensitivity of the document thresholds in the LCR algorithm, we employed the BM25 retriever combined with the RankGPT reranker, without the query threshold, and produced a heatmap illustrating NDCG@5 improvement percentages as a function of the lower threshold (LT) and upper threshold (UT). As shown in Figure~\ref{fig:param-dt}, ranking performance exhibits consistent enhancements across valid combinations of LT (0.1 to 0.9) and UT (0.2 to 1.0). Improvements intensify with higher UT values, peaking at UT=0.9. Gains are relatively modest when LT ranges from 0.5 to 0.7. Based on the heatmap, optimal configurations involve a high UT ($\sim$0.9) coupled with a low LT ($\sim$0.1--0.4). This approach enforces rigorous categorization: high-confidence documents advance to the top only under exceptionally strong confidence, low-confidence documents descend to the bottom only when confidence is notably weak, and medium-confidence cases preserve their initial positions amid uncertainty, thereby maximizing overall ranking effectiveness.

\subsection{Impact of Different Language Models}
\begin{figure}[pos=htbp]
\begin{tabular}{@{}c@{}c@{}}
\begin{subfigure}{0.5\linewidth}
\centering
\includegraphics[width=\linewidth]{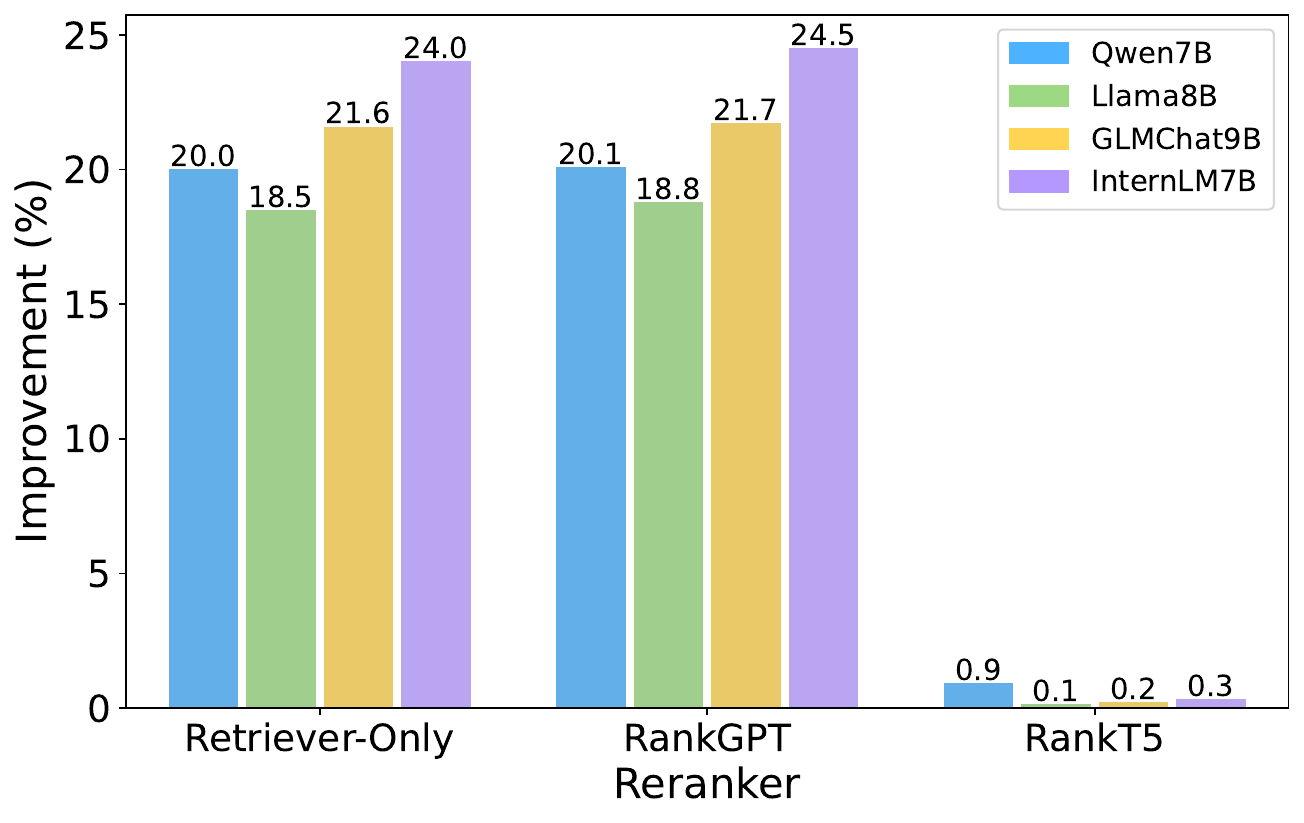}
\caption{BM25}
\label{fig:models-nq-bm25}
\end{subfigure} &
\begin{subfigure}{0.5\linewidth}
\centering
\includegraphics[width=\linewidth]{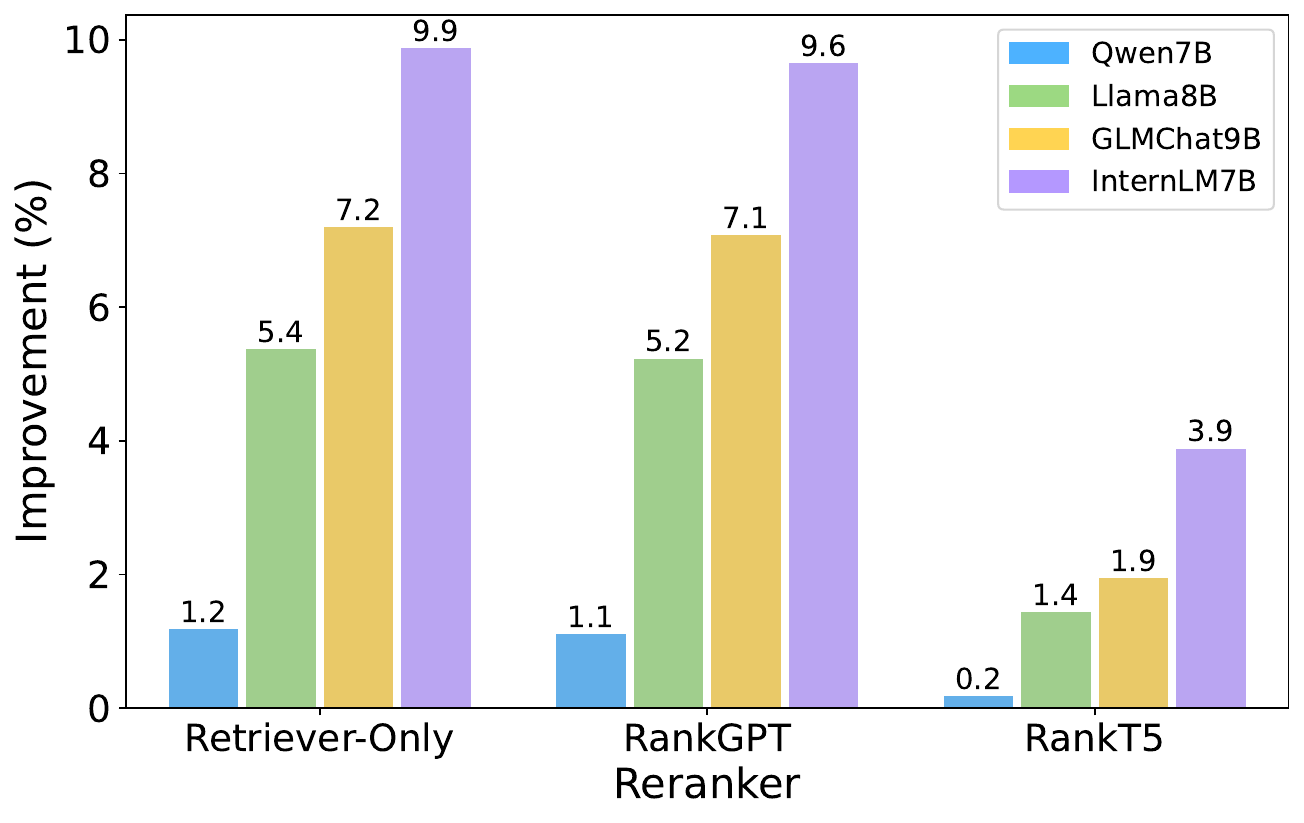}
\caption{Contriever}
\label{fig:models-nq-contriever}
\end{subfigure} \
\end{tabular}
\caption{\textbf{NDCG@5 Improvements Across Different Language Models.} Results are derived from the Natural Questions dataset.}
\label{fig:impact-of-models}
\end{figure}

To evaluate the effect of various large language models (LLMs) on the performance of the LCR algorithm, we tested four lightweight pre-trained models: Qwen2.5-7B-Instruct (Qwen7B)~\citep{yang2024qwen2}, Llama3.1-8B-Instruct (Llama8B)~\citep{grattafiori2024llama}, GLM-4-9B-Chat (GLMChat9B)~\citep{glm2024chatglm}, and InternLM2.5-7B-Chat (InternLM7B)~\citep{cai2024internlm2}. These models were selected due to their parameter sizes ranging from 7 to 9 billion, representing diverse architectures and training corpora, which allows us to assess the generalizability of LCR across different LLM families. We employed the MSCP method for confidence quantification. Experiments were conducted on the NaturalQuestions dataset, employing BM25 and Contriever as retrievers, consistent with the main experimental setup. As shown in Figure~\ref{fig:impact-of-models}, InternLM7B yields the highest NDCG@5 improvements across all configurations, such as 24.5\% for BM25 + RankGPT and 9.6\% for Contriever + RankGPT. GLMChat9B follows closely, while Qwen7B shows the least gains. These results indicate that model selection substantially influences confidence signal quality, with InternLM7B providing the strongest semantic understanding for reranking enhancement. Notably, all tested 7--9B-scale models consistently yield performance improvements, demonstrating that LCR is robust and generalizes well across specific model choices.

\subsection{Impact of Different Confidence Quantification Methods}

\begin{figure}[pos=htbp]
\begin{tabular}{@{}c@{}c@{}}
\begin{subfigure}{0.5\linewidth}
\centering
\includegraphics[width=\linewidth]{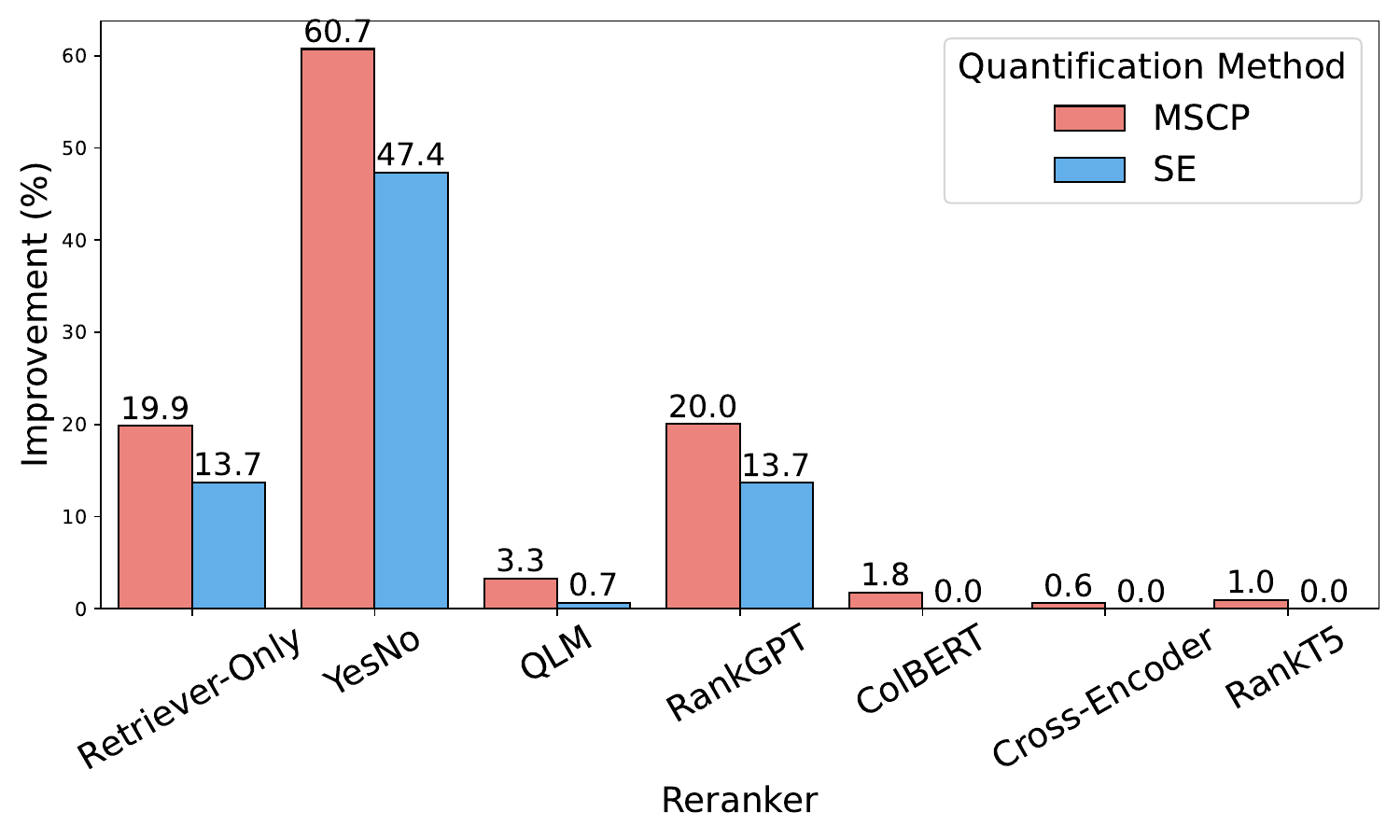}
\caption{BM25}
\label{fig:quant-method-nq-bm25}
\end{subfigure} &
\begin{subfigure}{0.5\linewidth}
\centering
\includegraphics[width=\linewidth]{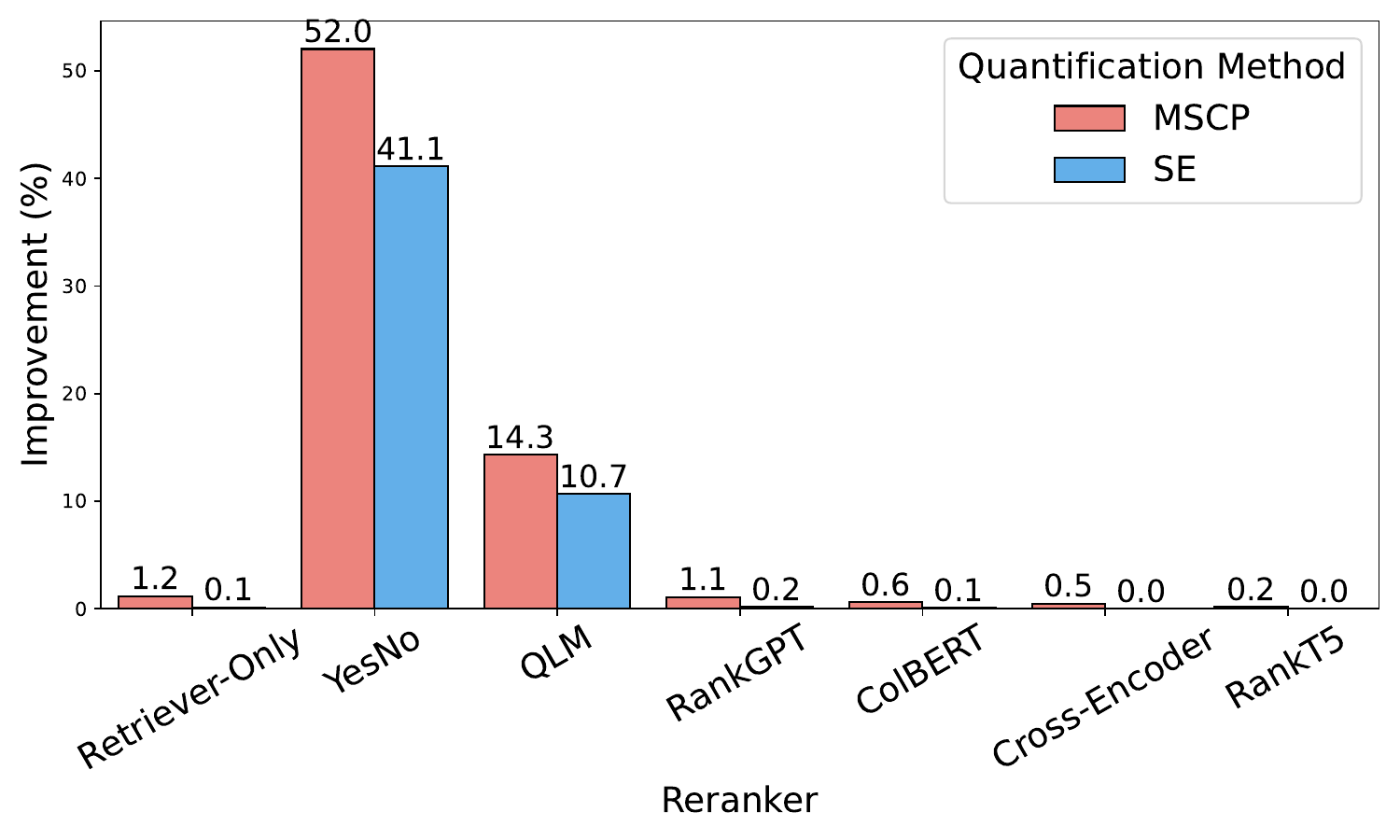}
\caption{Contriever}
\label{fig:quant-method-nq-contriever}
\end{subfigure} \\
\end{tabular}
\caption{\textbf{Impact of Confidence Quantification Methods on Performance Improvements.} Bar charts comparing NDCG@5 enhancements from MSCP and SE across rerankers, based on the NaturalQuestions dataset.}
\label{fig:impact-of-quant-methods}
\end{figure}

To examine the effects of diverse confidence quantification approaches on the LCR algorithm, we performed experiments on the NaturalQuestions dataset, utilizing BM25 and Contriever as initial retrievers. We integrated the LCR method with both Maximum Semantic Cluster Proportion (MSCP) and Semantic Entropy (SE)~\citep{farquhar2024detecting} applied to the outputs of various rerankers. Bar charts were generated to contrast the performance enhancements achieved by each method. As illustrated in Figure~\ref{fig:impact-of-quant-methods}, while SE yields improvements, it generally underperforms compared to MSCP. This disparity may arise because MSCP directly measures the dominance of the primary semantic cluster, thereby more robustly capturing semantic consistency, whereas SE, being an entropy-based metric, is more vulnerable to distributional noise in the sampled outputs.

\subsection{Underlying Mechanism}
\label{sec:mechanism}

\begin{figure}[pos=htbp]
\centering
\includegraphics[width=1.0\linewidth]{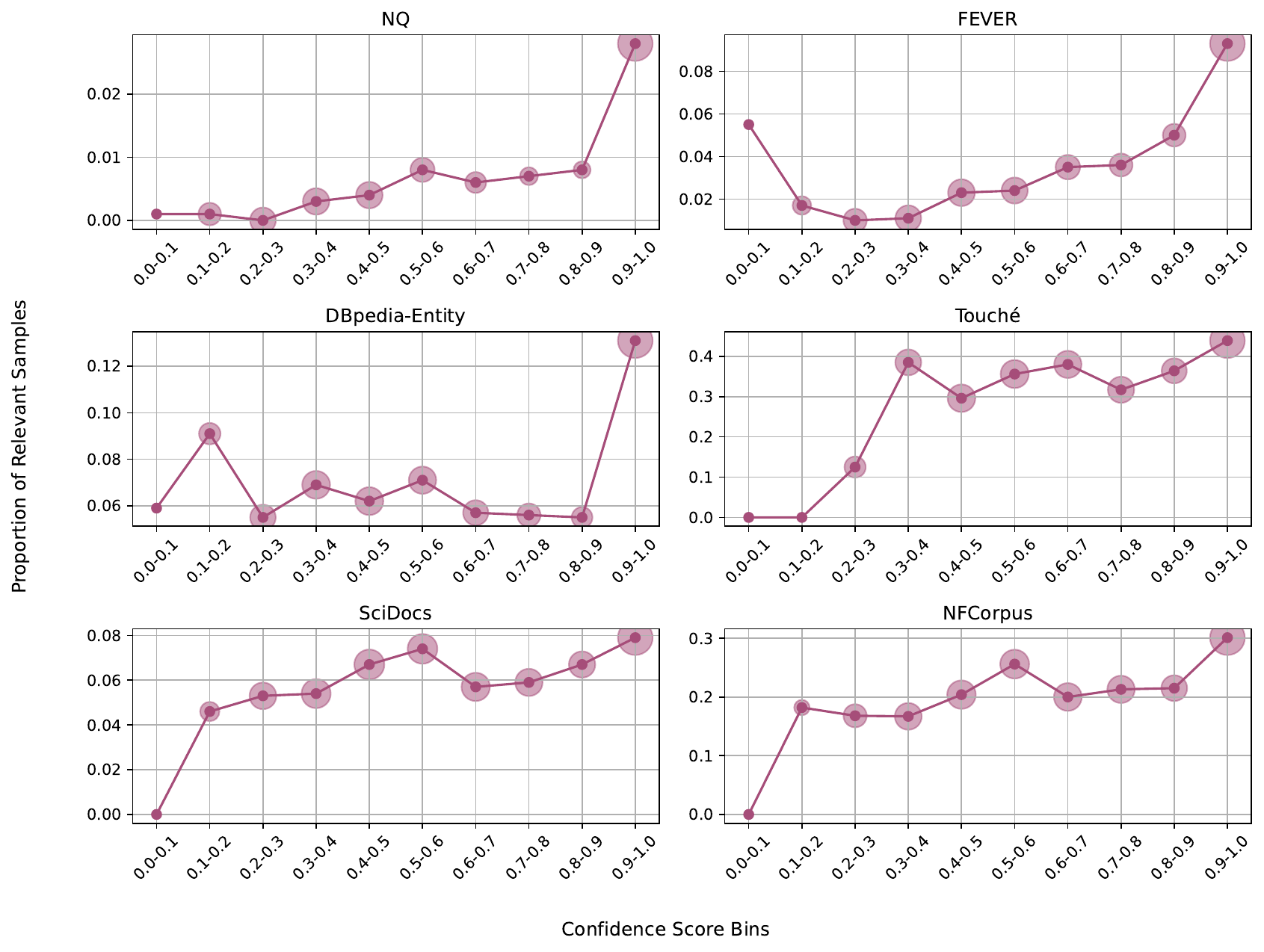}
\caption{\textbf{Confidence-Relevance Calibration Curve.} This figure depicts the relationship between confidence scores (derived from MSCP) and document relevance. Each subplot corresponds to a specific dataset, with confidence score bins along the x-axis and the proportion of relevant samples along the y-axis. The sizes of scatter points reflect the logarithmic sample percentage per bin, enhancing visibility amid varying distributions. Results are obtained using the BM25 retriever and the Qwen7B model.}
\label{fig:conf-rel}
\end{figure}

To elucidate the effectiveness of the LCR algorithm, we hypothesize that the confidence exhibited by LLMs in their generated responses positively correlates with the relevance of the input documents. This hypothesis is empirically validated through experiments conducted on six diverse datasets. We employ the MSCP for confidence quantification. For each query-document pair, confidence scores are computed and uniformly divided into 10 bins. Subsequently, we calculate the proportion of relevant documents within each bin to reveal the confidence-relevance association. To ensure consistent comparison, relevance scores are binarized across all datasets: scores greater than 0 are classified as relevant (1), while 0 denotes irrelevance (0). This binarization applies directly to datasets with binary relevance (NaturalQuestions, FEVER, SciDocs) and requires converting graded scores—treating values greater than 0 as relevant—for those with multi-level relevance (DBpedia-Entity, Touché, NFCorpus).
As illustrated in Figure~\ref{fig:conf-rel}, the majority of datasets demonstrate a clear positive correlation, wherein bins with higher confidence scores encompass larger proportions of relevant documents. This indicates that LLMs exhibit greater semantic consistency in responses informed by relevant documents, thereby generating more robust confidence signals that bolster LCR's ranking enhancements through confidence binning.

In summary, this empirical validation highlights the utility of LLMs' intrinsic semantic capabilities in document ranking, providing a theoretical foundation for the application of LCR in knowledge-intensive tasks.

\section{Conclusion}
\label{sec:conclusion}
In this paper, we addressed the persistent challenge of hallucinations in large language models (LLMs) for knowledge-intensive tasks by introducing the LLM-Confidence Reranker (LCR), a training-free, plug-and-play algorithm that harnesses LLMs' semantic understanding and question-answering capabilities to enhance document reranking in retrieval-augmented generation (RAG) systems. Unlike conventional rerankers that directly evaluate query-document relevance, LCR leverages black-box LLM confidence signals---quantified via the Maximum Semantic Cluster Proportion (MSCP)---as a proxy for relevance, employing a two-stage process of confidence assessment and multi-level sorting to prioritize relevant documents.

LCR demonstrates remarkable robustness across diverse settings, including various retrievers (sparse like BM25 and dense like Contriever), rerankers (pre-trained LLM-based and fine-tuned Transformer-based), and LLMs from different families. Relying solely on lightweight pre-trained models with 7--9B parameters, it ensures computational efficiency, pointwise independent scoring for parallelism, and scalability, all while maintaining black-box accessibility that enhances its practical utility in real-world deployments.

Comprehensive evaluations on BEIR and TREC benchmarks reveal consistent NDCG@5 enhancements of up to 20.6\% over baselines, without any degradation. Ablation studies elucidate the influence of key factors, and empirical validations affirm the hypothesis that LLM confidence positively correlates with document relevance, furnishing a robust theoretical foundation for LCR's efficacy.

By delivering strong generalization, minimal computational overhead, and effective hallucination mitigation, LCR significantly advances RAG systems. Future directions may explore adaptive thresholding, multi-LLM ensembles, and fusions with cutting-edge uncertainty quantification techniques to further elevate retrieval precision and overall RAG performance.

\clearpage
\bibliographystyle{apacite} 
\bibliography{reference}

\end{document}